\documentclass[10pt,twocolumn,letterpaper]{article}

\usepackage[preprint]{cvpr}      

\definecolor{cvprblue}{rgb}{0.21,0.49,0.74}
\usepackage[pagebackref,breaklinks,colorlinks,allcolors=cvprblue]{hyperref}
\usepackage[table,dvipsnames]{xcolor} 
\usepackage[most]{tcolorbox}
\usepackage{tabularx}
\usepackage{colortbl}    
\definecolor{lightgray}{gray}{0.96} 
\usepackage[normalem]{ulem} 
\usepackage{booktabs}
\usepackage[ruled,vlined,linesnumbered]{algorithm2e}
\usepackage{adjustbox} 
\usepackage[table]{xcolor} 
\usepackage{multirow}

\definecolor{lightblue}{RGB}{200, 220, 255}
\definecolor{midblue}{RGB}{140, 180, 250}
\definecolor{darkblue}{RGB}{90, 130, 230}
\definecolor{darkred}{RGB}{220,20,60}

\title{Where Culture Fades: \\ Revealing the Cultural Gap in Text-to-Image Generation}

\author{
Chuancheng Shi$^{1}$\thanks{Equal first authors.}  \quad
Shangze Li$^{2}$\footnotemark[1] \quad
Shiming Guo$^{1}$\footnotemark[1] \quad
Simiao Xie$^{1}$\thanks{Equal second authors.} \quad
Wenhua Wu$^{1}$\footnotemark[2] \quad
Jingtong Dou$^{1}$\\[2pt]
Chao Wu$^{2}$ \quad
Canran Xiao$^{3}$ \quad
Cong Wang$^{4}$ \quad
Zifeng Cheng$^{4}$ \quad
Fei Shen$^{5}$\thanks{Corresponding author.} \quad
Tat-Seng Chua$^{5}$\\[4pt]
$^{1}$The University of Sydney \quad
$^{2}$Nanjing University of Science and Technology \quad
$^{3}$Central South University\\
$^{4}$Nanjing University \quad
$^{5}$National University of Singapore
}

\begin{document}
\maketitle

\begin{abstract}
Multilingual text-to-image (T2I) models have advanced rapidly in terms of visual realism and semantic alignment, and are now widely utilized. Yet outputs vary across cultural contexts: because language carries cultural connotations, images synthesized from multilingual prompts should preserve cross-lingual cultural consistency. 
We conduct a comprehensive analysis showing that current T2I models often produce culturally neutral or English-biased results under multilingual prompts.
Analyses of two representative models indicate that the issue stems not from missing cultural knowledge but from insufficient activation of culture-related representations. 
We propose a probing method that localizes culture-sensitive signals to a small set of neurons in a few fixed layers. Guided by this finding, we introduce two complementary alignment strategies: (1) inference-time cultural activation that amplifies the identified neurons without backbone fine-tuned; and (2) layer-targeted cultural enhancement that updates only culturally relevant layers. 
Experiments on our CultureBench demonstrate consistent improvements over strong baselines in cultural consistency while preserving fidelity and diversity.
\end{abstract}

\section{Introduction}

Ensuring cultural fairness and representation in generative AI~\cite{deng2025acquire,ruiz2023dreambooth,tumanyan2023plug,jang2024rethinking,hu2024anomalydiffusion,he2024diffusion,jia2024ssmg,li2024cosmicman,shi2024instantbooth} is essential for global accessibility and cultural diversity, aligning with the United Nations’ principles of inclusiveness and universality. Yet, when prompted in different languages, many state-of-the-art methods~\cite{achiam2023gpt,ye2024altdiffusion,labs2025flux1kontextflowmatching,podell2023sdxl,esser2024scaling} frequently produce culturally neutral or English-biased images, which weakens cross-lingual cultural correspondence. Here, we use cultural consistency to denote the extent to which generated images exhibit visual elements that are statistically associated with the target language’s cultural context, beyond mere semantic correctness. To avoid conflating culturally typical elements with stereotypes, we require that cultural cues be both contextually appropriate and roughly consistent with real-world cultural statistics.

\begin{figure}[t]
\centering
\includegraphics[width=1\linewidth]{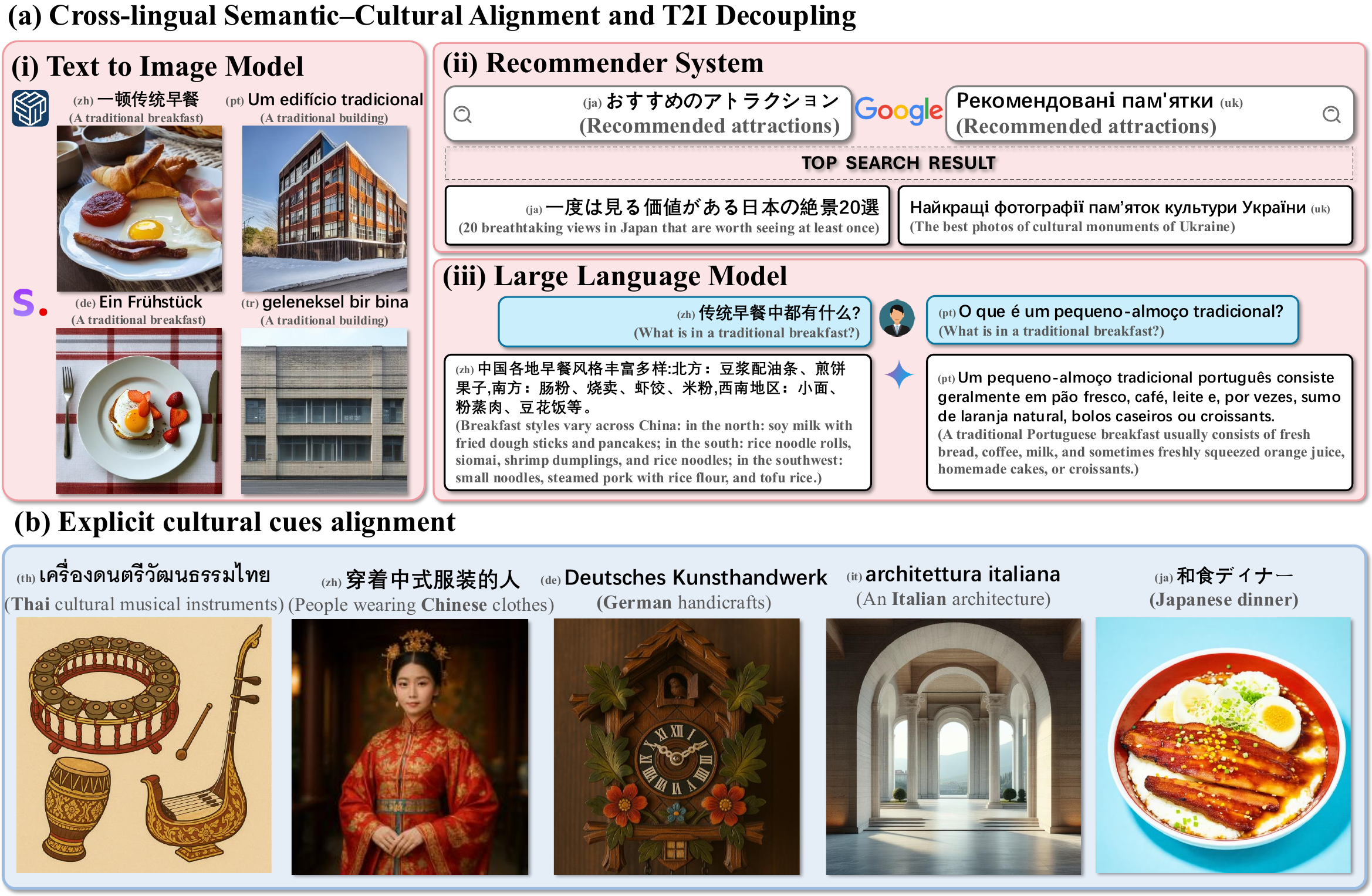}
\caption{\textbf{Cultural alignment in local languages.} (a) LLMs/recommenders keep cultural consistency, but T2I models falter with ``noun-only’’ prompts. (b) Adding a ``culture-style modifier + noun’’ restores consistency.}
\label{fig:q1}
\vspace{-0.5cm}
\end{figure}

\begin{figure*}[t]
\centering
\includegraphics[width=1\linewidth]{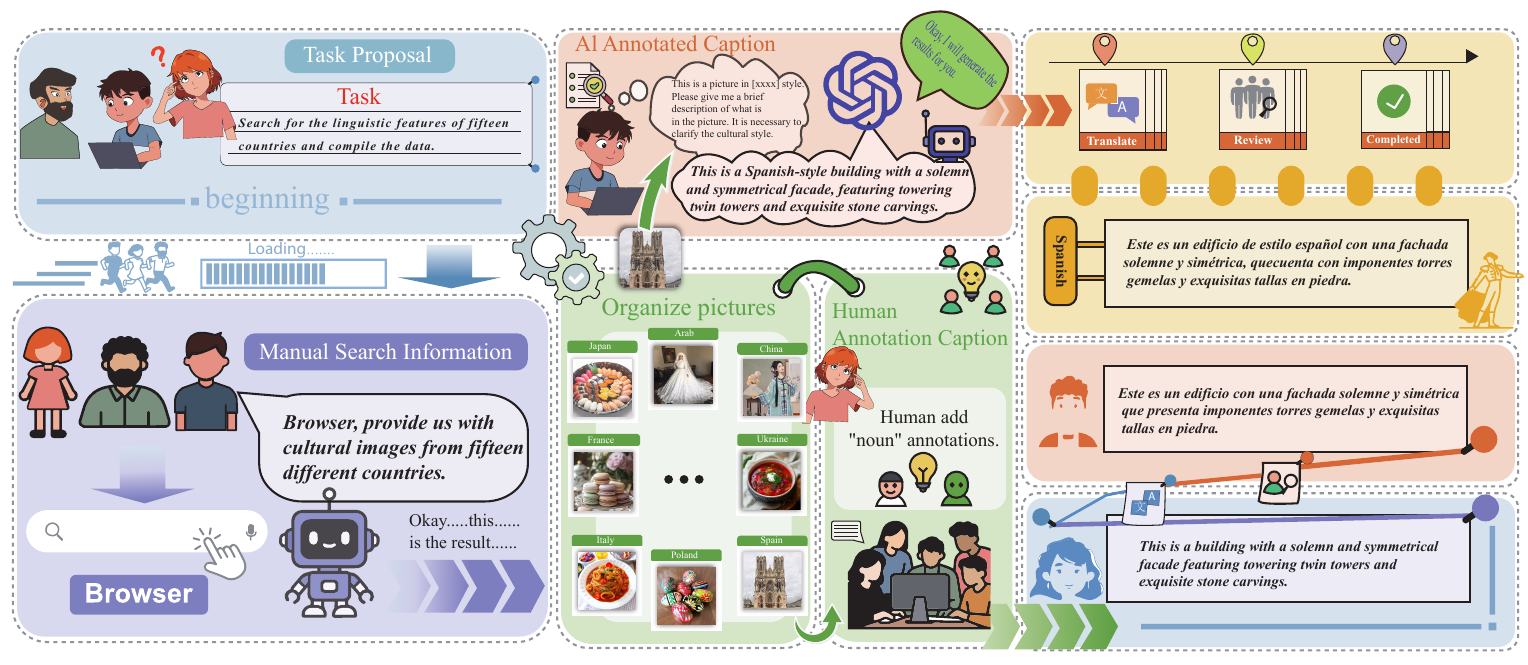}
\caption{\textbf{Overview of the CultureBench pipeline.} First, manually collect and rigorously quality-control datasets from 15 linguistic regions; annotate “culture-style modifier noun" captions using GPT5-Nano~\cite{chatgpt5} and, through human annotation, “noun-only" captions; convert annotated content into local languages via translation tools, supplemented by manual review.}
\vspace{-0.5cm}
\label{fig:data_pipe}
\end{figure*}

Despite strong gains in semantic and visual fidelity, diffusion-based T2I models~\cite{ramesh2022hierarchical, saharia2022photorealistic} lag in cultural generalization. Prior work largely targets cross-lingual encoder alignment~\cite{ma2024pea, ye2024altdiffusion} while overlooking cross-cultural grounding, defined as generating visuals that reflect each language’s socio-cultural context. As shown in Figure~\ref{fig:q1}(a), models often capture only literal meanings, for example, “\emph{a traditional building}” in Portuguese or Turkish, while missing culture-specific cues. In contrast, language-based systems such as LLMs~\cite{tang2024language, chatgpt5, achiam2023gpt} and recommender engines~\cite{ozsoy2024multilingual, iana2024mind} produce localized responses, revealing a cross-modal gap in cultural grounding.

We argue that the issue is insufficient activation rather than missing knowledge. Large-scale training corpora already contain diverse cultural attributes, and explicit prompts can elicit them. As shown in Figure~\ref{fig:q1}(b), adding culture-style modifiers, for example, “\emph{people wearing Chinese clothes}” or “\emph{an Italian architecture},” yields images with clear country-specific characteristics. The inconsistency happens because ``noun-only” prompts don’t strongly trigger cultural knowledge, so the model gives literal but not cultural interpretations.

To validate this hypothesis and mitigate cultural inconsistency in T2I models, we first examine two representative systems~\cite{ye2024altdiffusion, ma2024pea} and show that the failure arises from insufficient activation of culture-related knowledge rather than its absence, and that the effect is observed across two architecturally different diffusion methods. We then introduce a two-stage probing method. We begin by comparing attention distributions between culture-style modifiers and nouns to localize culture-sensitive layers. Next, we use the Top-K SAE~\cite{cunningham2023sparse} to quantify activation differences between prompts with explicit cultural cues and ``noun-only’’ prompts, revealing that culture-relevant representations cluster in a few fixed layers and a small set of neurons. Guided by these findings, we propose two complementary strategies: (1) a zero-training activation scheme that amplifies the responses of the identified neurons at inference time, and (2) a layer-specific fine-tuned scheme that updates only culturally relevant layers to improve consistency. Finally, experiments on our CultureBench show consistent improvements over strong baselines in cultural consistency while preserving fidelity and diversity.
We highlight the following contributions:
\begin{itemize}
  \item We empirically show that multilingual T2I models often produce culturally neutral or subtly English-biased outputs under multilingual prompts, thereby hindering cross-lingual cultural consistency.
  
  \item We develop a probing framework that localizes culture-sensitive signals to a few fixed layers and neurons by contrasting attention patterns and Top-K SAE activations, indicating that failures stem from insufficient activation rather than missing knowledge.

  \item We propose two alignment strategies: a zero-training inference-time activation scheme and a layer-specific fine-tuning scheme that updates only culturally relevant layers, improving cultural consistency while preserving fidelity and diversity.

  \item We curate and release CultureBench, a 15-country benchmark with multilingual prompts and images, enabling evaluation of cross-cultural consistency and culturally adaptive training.
\end{itemize}

\section{Related Work}

\noindent\textbf{Cultural Text-to-Image Generation.}
Prior work has examined cultural bias, fairness, and stereotyping in T2I systems~\cite{friedrich2023fair,luccioni2023stable,jeong2025culture}. For example, SCoFT~\cite{Liu_2024_CVPR} expands cultural coverage through CCUB to mitigate stereotyping, and ViSAGe~\cite{jha2024visage} quantifies visual stereotypes. However, existing studies remain largely focused on fairness and debiasing, with a strong English-centric orientation and limited coverage of low-resource languages and diverse cultural contexts~\cite{d2024openbias, seshadri2024bias}. Clear definitions and systematic evaluation of cross-lingual cultural consistency are still lacking. Noun-only or short prompts often collapse to culturally neutral or implicitly English-biased generations in multilingual settings, yet a unified diagnostic framework is absent~\cite{li2023translation}. At the representational level, prior work provides little insight into where culture-sensitive features reside within the model or how to control them at the layer or neuron level~\cite{kumari2023ablating,orgad2023editing}. Methodologically, there is still a lack of lightweight, plug-and-play interventions and corresponding benchmarks that do not require large-scale retraining.

\begin{figure}[t]
\centering
\includegraphics[width=0.9\linewidth]{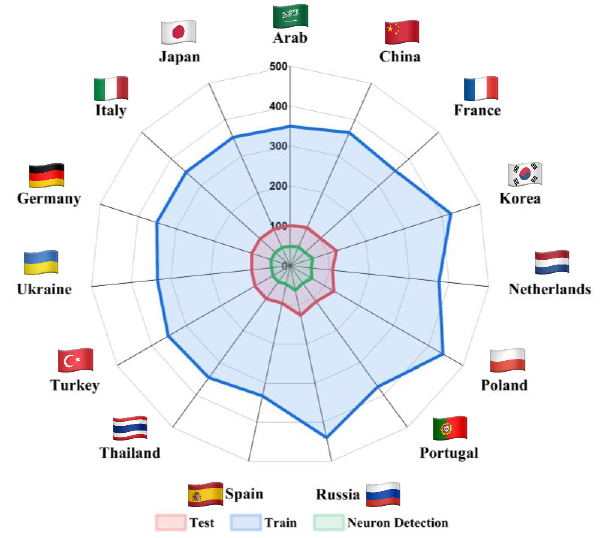}
\caption{\textbf{Data distribution of the proposed CultureBench dataset across 15 languages.} The dataset is divided into train, test, and neuron-detection subsets with a ratio of 7:2:1.}
\label{fig:dataset}
\vspace{-0.3cm}
\end{figure}

\noindent\textbf{Neuron Interpretability.}
Neuron-level interpretability links model activations to learned concepts in vision and language~\cite{mu2020compositional,huo2024mmneuron,chen2025identifying,yu2024interpreting}. Recent work further distinguishes shared and language-specific semantics at the neuron or direction level, clarifying how abstract meanings distribute across layers~\cite{huang2025neurons,tang2024language,xie2021importance}. In multimodal settings, causal probing shows that individual neurons can be driven by visual concepts via textual proxies and directly manipulated to control semantics~\cite{schwettmann2023multimodal,pan2023finding,he2025single}. For concreteness, FEMN~\cite{pan2023finding} locates sparse neuron groups mediating specific concepts (e.g., “smile,” “striped”) in a CLIP-style model and demonstrates that directly patching or amplifying those units causally increases or suppresses the target concept in zero-shot classification and retrieval, establishing neuron-level semantic control. However, prior studies still focus mainly on object- or attribute-level notions and provide limited diagnostics for cultural representations, especially those that vary under multilingual prompts~\cite{mitchell2025shades,fort2024your,wang2024seaeval}.

\section{CultureBench Dataset}

\noindent\textbf{Data Collection.}
As shown in Figure~\ref{fig:data_pipe}, we collect culturally representative images across 15 language and region groups via geo-constrained web search using native and translated keywords. Each image is manually verified for clarity, authenticity, and representativeness to ensure a faithful depiction of its target culture.

\noindent\textbf{Data Annotations.} CultureBench provides multidimensional annotations for assessing cultural awareness, including cultural categories, geographic regions, languages, and image content. For each sample, we provide two textual descriptions: a “culture-style modifier + noun’’ caption generated by GPT5-Nano~\cite{chatgpt5} and a human-written “noun-only’’ caption. To reduce subjectivity in defining and labeling cultural attributes, we invited domain experts to review the taxonomy and a subset of samples. Annotators kept only culturally appropriate, statistically grounded cues and treated mismatched ones as stereotypes.

\begin{figure}[t]
\centering
\includegraphics[width=0.8\linewidth]{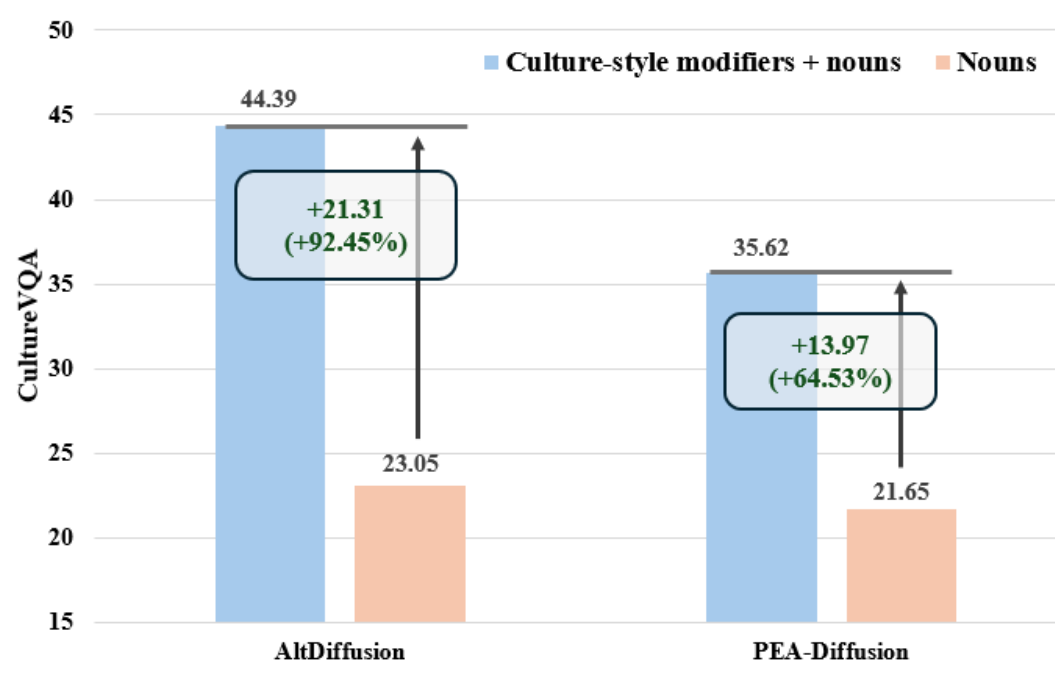}
\caption{\textbf{Verify the hypothesis.} Within the CultureBench test subset, performances under “culture-style modifier + noun” and “noun-only” prompt conditions are compared. Quantitative evaluation is conducted using CultureVQA.}
\vspace{-0.3cm}
\label{fig:analy}
\end{figure}

\noindent\textbf{Data Distribution.} 
As shown in Figure~\ref{fig:dataset}, the dataset contains 7{,}932 samples split into train, test, and neuron-detection subsets at a 7:2:1 ratio: the train subset is used for model and adapter training, the test subset for quantitative evaluation, and the neuron-detection subset for layer- and neuron-level probing. Training is strictly limited to the train subset. Neither the test subset nor the neuron-detection subset is accessed during training, hyperparameter tuning, or model selection.

\noindent\textbf{Cultural Evaluation.}
We propose CultureVQA, a single-choice VQA task built on CultureBench. For each image, Qwen3-VL~\cite{yang2025qwen3} and Gemini-2.5-Flash~\cite{comanici2025gemini} must choose one label from 15 language/region categories or an “unrecognisable” option. We report accuracy as the proportion of samples whose predicted label exactly matches the ground truth. In a pilot study, we found that the cultural labels predicted by these models are highly consistent with human annotations, supporting their use as automatic evaluators. This setup tests a model’s ability to perform cultural attribution from visual cues alone, without textual prompts.

\section{Cultural Probing}

\begin{figure*}[t]
\centering
\includegraphics[width=0.95\linewidth]{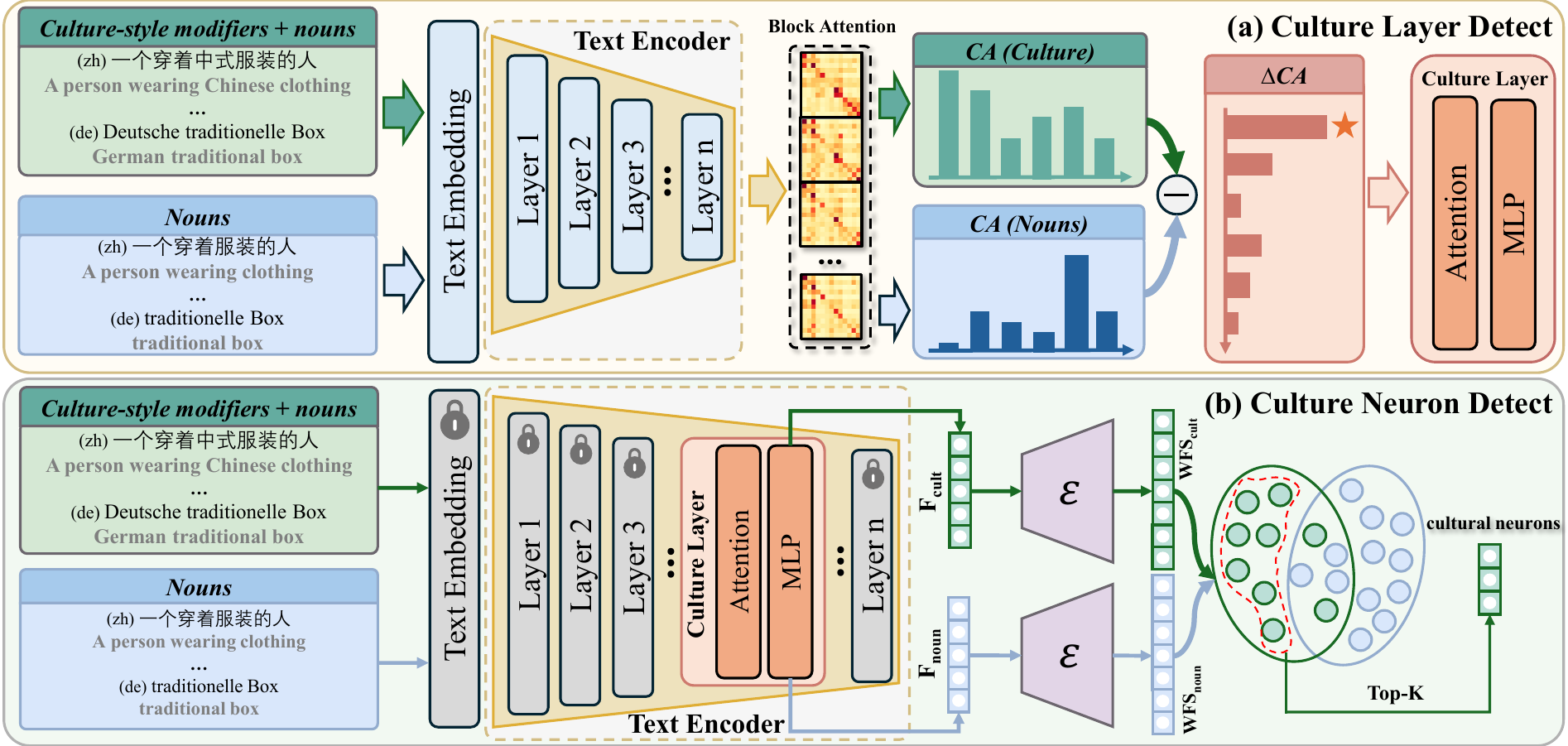}
\caption{\textbf{Methods for Neuronal Detection.} (a) By comparing attention allocation between cultural-style modifiers and nouns across text-encoder layers, the layer with the largest divergence is designated as the culturally sensitive layer. (b) At this layer, features from the “culture-style modifier + noun’’ and “noun-only’’ prompts are fed into an SAE~\cite{cunningham2023sparse} to obtain sparse features, revealing neurons with heightened sensitivity to cultural cues.}
\label{fig:detect}
\vspace{-0.5cm}
\end{figure*}

\subsection{Cultural Gap}
To test the hypothesis that the issue lies in insufficient activation rather than knowledge absence, we conducted a unified controlled experiment. Specifically, we used prompts composed of “culture-style modifier + noun” as inputs to AltDiffusion~\cite{ye2024altdiffusion} and PEA-Diffusion~\cite{ma2024pea}, generated corresponding images, and evaluated them using the CultureVQA accuracy metric. As a control, we repeated the same process with ``noun-only’’ prompts. As shown in Figure~\ref{fig:analy}, prompts combining “culture-style modifier + nouns” achieved the best CultureVQA performance, 44.39 for AltDiffusion and 35.62 for PEA-Diffusion, both substantially higher than the ``noun-only’’ prompts. This consistent trend across two structurally different diffusion models suggests that the phenomenon is not model-specific and provides empirical support for our hypothesis.

\subsection{Culture Layer Detection}
As illustrated in Figure~\ref{fig:detect}(a), we compare the attention directed towards the target noun under paired “culture-style modifier + noun” and “noun-only” prompts, averaging across heads to obtain hierarchical cultural relevance scores. “Culture-style modifier + noun” prompts consistently yield significantly higher scores than ``noun-only’’ prompts, indicating that they are encoded with cultural semantics.

We define paired prompts for each target concept: a “culture-style modifier + noun” prompt \(P_{\text{cult}}\) and a ``noun-only’’ prompt \(P_{\text{noun}}\). The former augments the noun with a culture-style modifier, while the latter removes it. We create $N$ such pairs covering diverse cultural elements to ensure generality.

For each prompt, we annotate two token categories, the culture modifiers \(T_{\text{cult}}\) and the target nouns \(T_{\text{noun}}\). And extract layerwise attention. Let $A(l) \in \mathbb{R}^{B \times H \times S \times S}$ be the multi-head attention at layer $l$, where \(B\) is the batch size, \(H\) is the number of attention heads, and \(S\) is the sequence length. We average heads for robustness: 
\vspace{-0.25cm}
\begin{equation} 
\bar{A}{(l)} = \frac{1}{H}\sum_{h=1}^{H} A_{h}(l), 
\end{equation} 
and retain only the entries from \(T_{\text{cult}}\) to \(T_{\text{noun}}\). Subsequently, we derive the subset of keyword pairs from \(\bar{A}{(l)}\) denoted as \(\bar{A}_{\text{key}}{(l)}\in \mathbb{R}^{B \times \lvert T_{\text{cult}}\rvert \times \lvert T_{\text{noun}}\rvert}\).

\begin{figure}[t]
\centering
\includegraphics[width=0.95\linewidth]{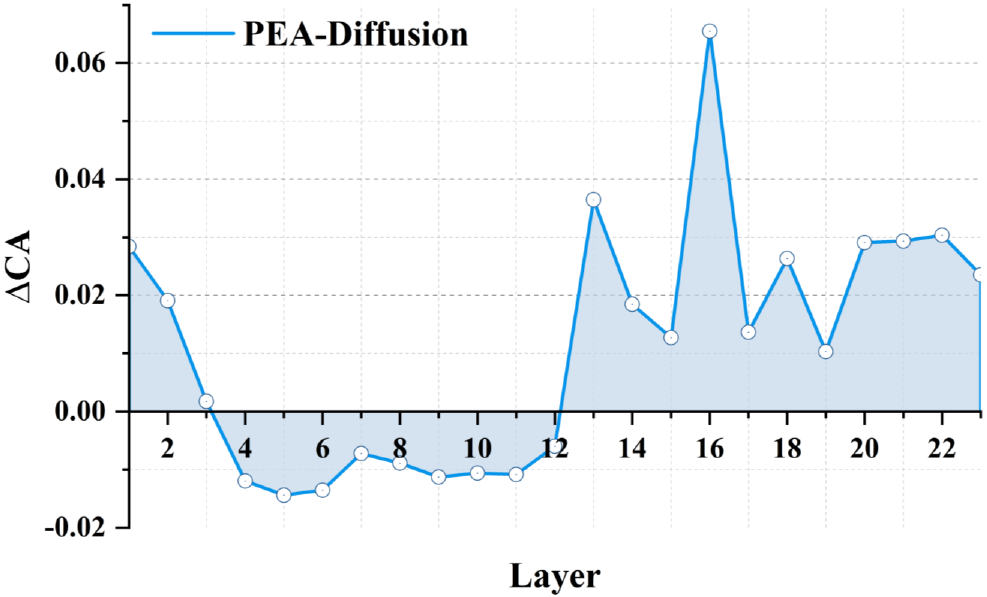}
\vspace{-0.3cm}
\caption{\textbf{PEA-Diffusion cultural sensitivity.} $\Delta$CA peaks layer 16. AltDiffusion results are provided in the appendix.}
\vspace{-0.5cm}
\label{fig:layer}
\end{figure}
    
If layer \(l\) encodes culture semantics, the attention from culture modifiers to target nouns under a cultural prompt should exceed that under its ``noun-only’’ prompt. To quantify this, let $t_{\text{cult}} \in T_{\text{cult}}$ and $t_{\text{noun}} \in T_{\text{noun}}$ denote a cultural modifier and target noun. For a prompt \(P\), the attention from cultural modifiers to target nouns in layer \(l\) is:
\vspace{-0.3cm}
\begin{equation}
\operatorname{CA}(P,l) = \frac{\sum_{t_{\text{cult}} } \sum_{t_{\text{noun}}} \bar{A}_{key}{(l)}_{\,t_{\text{cult}} \to t_{\text{noun}}}}{\lvert T_{\text{cult}}\rvert \cdot \lvert T_{\text{noun}}\rvert}.
\end{equation}
For all culture/noun pairs \((P_{\text{cult}},P_{\text{noun}})\), we compute
\begin{equation}
\Delta \text{CA}(l) = \frac{1}{N} \sum_{i=1}^{N} [\text{CA}(P_{\text{cult},i}, l) - \text{CA}(P_{\text{noun},i}, l)].
\end{equation}
A Larger \(\Delta\operatorname{CA}(l)\) indicates that layer \(l\) more effectively separates “culture-style modifier’’ from “noun’’ semantics.

We compute \(\Delta\operatorname{CA}(l)\) across prompt pairs and seeds, and mark a layer as culture-sensitive if its \(\Delta\operatorname{CA}\) notably exceeds the mean of its two neighboring layers.
Based on this workflow, Figure~\ref{fig:layer} shows the detection results from PEA-Diffusion, revealing clear global peaks at a specific layer.

\begin{tcolorbox}[breakable, colback=Goldenrod!30, colframe=Bittersweet!80,
                  boxrule=0.5pt, arc=3pt, left=3pt, right=3pt,
                  top=2pt, bottom=2pt]
The results indicate that culturally relevant semantics are not uniformly distributed throughout the network, but rather concentrated in a key layer.
\end{tcolorbox}

\subsection{Culture Neuron Detection}
We localize culturally sensitive neurons in the key layer. As shown in Figure.~\ref{fig:detect}(b), we apply a Top-K SAE~\cite{cunningham2023sparse} to obtain a decomposition of attention features and select culture-specific neurons via comparative analysis.

From the key layer, we construct cultural features \(F_{\text{cult}} \in \mathbb{R}^{N_{\text{cult}} \times D_{\text{att}}}\) and noun features \(F_{\text{noun}} \in \mathbb{R}^{N_{\text{noun}} \times D_{\text{att}}}\), where \(N_{\text{cult}}\) and \(N_{\text{noun}}\) are the numbers of token pairs for the cultural and noun prompts, respectively. The attention feature dimension is \(D_{\text{att}} = |T_{\text{cult}}| \times |T_{\text{noun}}|\).
\begin{figure}[t]
\centering
\includegraphics[width=0.9\linewidth]{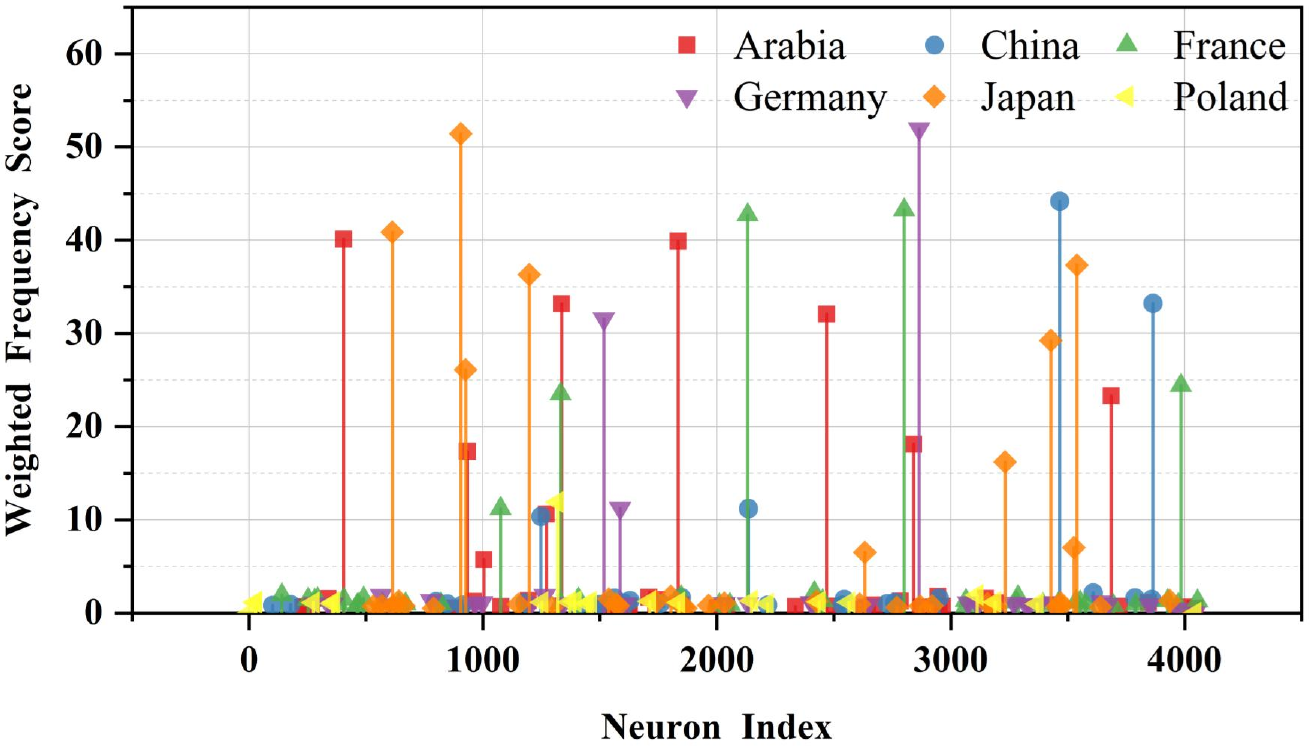}
\vspace{-0.3cm}
\caption{\textbf{Neuronal detection result}. The weighted frequency scores show only a few salient peaks per culture, indicating culture-specific neurons. We define the Top-$K$ set as the peak neurons, with $K$ adapting to the number of salient peaks.}
\vspace{-0.4cm}
\label{fig:neuron}
\end{figure}
We use a weighted frequency score combining activation frequency and amplitude to quantify neuronal responses to cultural semantics. Specifically, let \(m \in \{1, 2, \dots, D_{\text{att}}\}\) indexes a specific neuron in the attention feature space. The activation frequency $f_{\text{cult}}(m)$ is defined as follows: for all $N_{\text{cult}}$ lexemes with cultural elements, compute the proportion of samples in which the activation $Z_{\text{cult}}[i,m]$ of the $m$th neuro exceeds $\epsilon$:
\begin{equation}
    f_{\text{cult}}(m) = \frac{1}{N_{\text{cult}}} \sum_{i=1}^{N_{\text{cult}}} \mathbb{I}\left( Z_{\text{cult}}[i, m] > \epsilon \right).
\end{equation}
Here $i$ denotes the $i$th sample in $N_{\text{cult}}$, $ \mathbb{I}(\cdot)$ is the indicator function.Using the same method, we obtained $f_{\text{noun}}(m)$. 

For each neuron, we define the mean activation magnitude $\mu_{\text{cult}}(m)$ by calculating the magnitude of its activation in the cultural attention feature and adding it to the magnitude of its activation in the noun attention feature token:
\begin{equation}
\mu_{\text{cult}}(m) = \frac{\sum_{i=1}^{N_{\text{cult}}} \left( Z_{\text{cult}}[i, m] \cdot \mathbb{I}\left( Z_{\text{cult}}[i, m] > \epsilon \right) \right)}{\sum_{i=1}^{N_{\text{cult}}} \mathbb{I}\left( Z_{\text{cult}}[i, m] > \epsilon \right) + \beta}.
\end{equation}
Here $\beta$ denotes a small positive constant added for numerical stability, ensuring that the denominator never becomes zero and preventing degenerate cases when the activation frequency is extremely low. Using the same method, we obtained $\mu_{\text{noun}}(m)$. $\mu_{\text{noun}}(m)$ denotes the average activation magnitude of the $m$th neuron on the noun subset $F_{\text{noun}}$.

Ultimately, to better capture neurons with both high firing rates and strong responses, we combine the activation frequency of cultural modifiers with their average activation magnitude to obtain a weighted frequency score $WFS_{\text{cult}}$:
\begin{equation}
WFS_{\text{cult}}(m)=f_{\text{cult}}(m) \cdot \mu_{\text{cult}}(m).
\end{equation}
Using the same method, we obtained $WFS_{\text{noun}}(m)$. $WFS_{\text{noun}}(m)$ is the weighted frequency score for noun subsets. After computing $WFS_{\text{cult}}$ and $WFS_{\text{noun}}$, we rank neurons by $WFS_{\text{cult}}$ and select the top-$K$ candidates. Neurons with substantial noun-side salience are removed, and the remaining ones are designated as culturally sensitive neurons.

According to the aforementioned algorithm, as illustrated in Figure~\ref{fig:neuron}, six distinctly different cultural backgrounds are presented; in each instance, one or more pronounced peaks emerge, each bearing a different index value.

\begin{tcolorbox}[breakable, colback=Goldenrod!30, colframe=Bittersweet!80,
                  boxrule=0.5pt, arc=3pt, left=3pt, right=3pt,
                  top=2pt, bottom=2pt]
This indicates that the locations of corresponding cultural neurons do not overlap across cultures.
\end{tcolorbox}

\renewcommand{\arraystretch}{1}
\begin{table}[t]
\centering
\caption{\textbf{Validating Culture-Sensitive Neuron Detection in PEA-Diffusion.} Neuronal accuracy on the test subset with “cultural style modifier + noun” prompts.}
\vspace{-0.2cm}
\label{tab:alt_cacc_ablation}
\adjustbox{max width=\linewidth}{
\begin{tabular}{l|c}

\hline
Method & CultureVQA $\uparrow$ \\
\hline
\rowcolor{gray!20}PEA-Diffusion~\cite{ma2024pea} & 35.62 \\
+ Masked Top-K Neurons & 7.65 \scriptsize{\color{darkred}{(-27.97)}} \\
+ Masked Random Neurons & 33.04 \scriptsize{\color{darkred}{(-2.58)}} \\
\hline
\end{tabular}
}
\vspace{-0.3cm}
\end{table}

\subsection{Cultural Validation}
To verify whether our detector accurately localizes culture-sensitive neurons, we designed three controlled settings (Table~\ref{tab:alt_cacc_ablation}): (1) Baseline (no masking), (2) Masked Top-K Neurons (masking the Top-K culture-sensitive neurons identified by our method), and (3) Random Mask (masking the same number of neurons at random). Masking the identified Top-K neurons reduces the mean CultureVQA score from 35.62 to 7.65, whereas random masking yields a comparable score (33.04) to the baseline. The sharp, targeted degradation absent under random masking indicates that the localized neurons are highly related to cultural semantics, thereby providing empirical support for the accuracy and interpretability of our neuron localization method.

\section{Methods}

\subsection{Zero-Training Neuron Amplifier}
Upon identifying neurons sensitive to cultural information, we amplify underutilised cultural representations by intervening on a targeted neuronal subset \(M_{\mathrm{cult}}\) within the key layer, thereby modulating attention-related features and enhancing cultural attributes in the generated images.

We define the original attention correlation features that need to be interfered \ $F_{\text{raw}} \in \mathbb{R}^{B \times S_{\text{pair}} \times D_{\text{att}}}$ \, where $B$ is the batch size, $S_{pair}$ is the number of prompt words in a single batch. The latent vector $Z_{\text{raw}} \in \mathbb{R}^{B \times S_{\text{pair}} \times M_{cult}}$ after inputting them into the SAE encoder is given by:
\begin{equation}
    Z_{\text{raw}} = \text{SAE.encode}(F_{\text{raw}}), 
\end{equation}
where the SAE, through sparse coding, decomposes the complex internal representations into more independent and semantically coherent neurons. At the same time, $Z_{\text{raw}}[b,p,m]$ denotes the initial activation value of the $m$-th neuron for the $p$-th prompt word in the $b$-th batch.
We enhance cultural representation by modulating culturally specific neurons with manually defined $\lambda$. The formula is:
\begin{equation}
    Z_{\text{enh}}[b, p, m] = 
    \begin{cases} 
    (1+\lambda) Z_{\text{raw}}[b, p, m]  & \text{if } m \in M_{\text{cult}} \\
    Z_{\text{raw}}[b, p, m] & \text{otherwise}
    \end{cases}.
\end{equation}
Here $Z_{\text{enh}} \in \mathbb{R}^{B \times S_{\text{pair}} \times M}$ \ is the modulated latent vector , $M$ is the collection of all neurons. $\lambda \in \mathbb{R}$ is the feature fusion coefficient. We then map $Z_{enh}$ back to the attention space:
\begin{equation}
    F_{\text{rec\_enh}} = \text{SAE.decode}(Z_{\text{enh}}),
\end{equation}
where $\text{SAE.decode}(\cdot)$ represents the SAE decoder and $F_{\text{rec\_enh}} \in \mathbb{R}^{B \times S_{\text{pair}} \times D_{\text{att}}}$ \ is the attention association feature reconstructed after modulation. This step preserves the original semantic structure while enhancing culturally specific attention patterns.

\subsection{Fine-Tuned Layer Enhancer}
To enable adaptive cultural representation without the need for manual adjustment of modulation strength, we propose a layer-specific, fine-tuned scheme that updates only culturally relevant layers to improve consistency.
We first analyze the text encoder layer by layer to identify the layer most sensitive to cultural cues, denoted as $l_c$.
A small trainable module is inserted only into this layer, while all other parameters remain frozen.

Let $h$ denote the hidden representation of the text encoder at layer $l_c$.
The enhancer produces an enhanced representation $\tilde{h}$ via a residual transformation:
\begin{equation}
    \tilde{h} = h + g\!\big(W_2\,\sigma(W_1 h)\big),
\end{equation}
where $\tilde{h}$ is the culture-enhanced hidden state, $\sigma(\cdot)$ is a nonlinear activation, $g(\cdot)$ denotes a normalization layer used to stabilize the residual transformation, and $W_1$, $W_2$ are small trainable matrices.
During training, only the enhancer parameters are updated while keeping the backbone fixed. Given a ``noun-only’’ prompt $p$, the text encoder with enhancer $f_{\theta,\phi}$ and generator $G$ produce an image:
\begin{equation}
    \hat{x} = G\!\big(f_{\theta,\phi}(p)\big).
\end{equation}
The generated image is compared with a ground-truth cultural image $x^{*}(p)$,
which is directly taken from the CultureBench dataset as the human-curated
cultural reference corresponding to the ``noun-only’’ prompt $p$, using a pixel-level
mean squared error (MSE) loss:
\begin{equation}
    \mathcal{L}_{\text{MSE}} = \frac{1}{N}\sum_{i=1}^{N}
    \big\|\hat{x}_{i} - x^{*}_{i}(p)\big\|_2^2,
\end{equation}
where $N$ is the number of pixels.
Only the enhancer parameters are optimized:
\begin{equation}
    \phi^{*} = \arg\min_{\phi}\, \mathcal{L}_{\text{MSE}}.
\end{equation}
During inference, the trained layer enhancer modulates the hidden representation $\tilde{h}$ according to the input prompt, thereby enhancing cultural consistency in the generated images while maintaining the original semantic structure.

\renewcommand{\arraystretch}{1}
\begin{table*}[t]
\centering
\caption{\textbf{Quantitative comparisons with SOTA methods.} Using the “noun-only'' prompts on the test subset. The best performance is marked in \textbf{bold}, and the second-best is \underline{underlined}.}
\vspace{-0.3cm}
\adjustbox{max width=\linewidth}{
\begin{tabular}{lcccc}
\hline
\textbf{Method} & \textbf{CultureVQA $\uparrow$} & \textbf{CLIPScore $\uparrow$} & \textbf{ImageReward $\uparrow$} & \textbf{LPIPS $\downarrow$} \\
\hline
StableDiffusion XL~\cite{podell2023sdxl} & 9.36 & 0.211 & -1.82 & 0.756 \\
FLUX.1-dev~\cite{labs2025flux1kontextflowmatching} & 14.83 & 0.224 & -0.88 & 0.692 \\
Show-o2~\cite{xie2025show} & 16.43 & 0.234 & -0.91 & 0.691 \\
PEA-Diffusion~\cite{ma2024pea} & 21.65 & 0.253 & -0.65 & 0.673 \\
AltDiffusion~\cite{ye2024altdiffusion} & 23.05 & 0.282 & -0.11 & 0.688 \\
StableDiffusion 3.5~\cite{esser2024scaling} & 25.13 & 0.242 & -1.01 & 0.715 \\
\hline
\rowcolor{lightgray} \textbf{Ours (Zero-Training)} & \underline{33.91} \color{darkred}{\scriptsize{(+12.32)}} & \textbf{0.291} \color{darkred}{\scriptsize{}(+0.038)} & \textbf{0.33} \color{darkred}{\scriptsize{}(+0.98)} & \textbf{0.654} \color{ForestGreen}{\scriptsize{}(-0.019)} \\
\rowcolor{lightgray} \textbf{Ours (Fine-Tuned)} & \textbf{36.63} \color{darkred}{\scriptsize{}(+14.98)} & \underline{0.290} \color{darkred}{\scriptsize{}(+0.037)} & \underline{0.31} \color{darkred}{\scriptsize{}(+0.42)} & \underline{0.661} \color{ForestGreen}{\scriptsize{}(-0.012)} \\
\hline
\end{tabular}
\label{tab:quantitative}
}
\end{table*}

\begin{figure*}[t]
\centering
\includegraphics[width=0.95\linewidth]{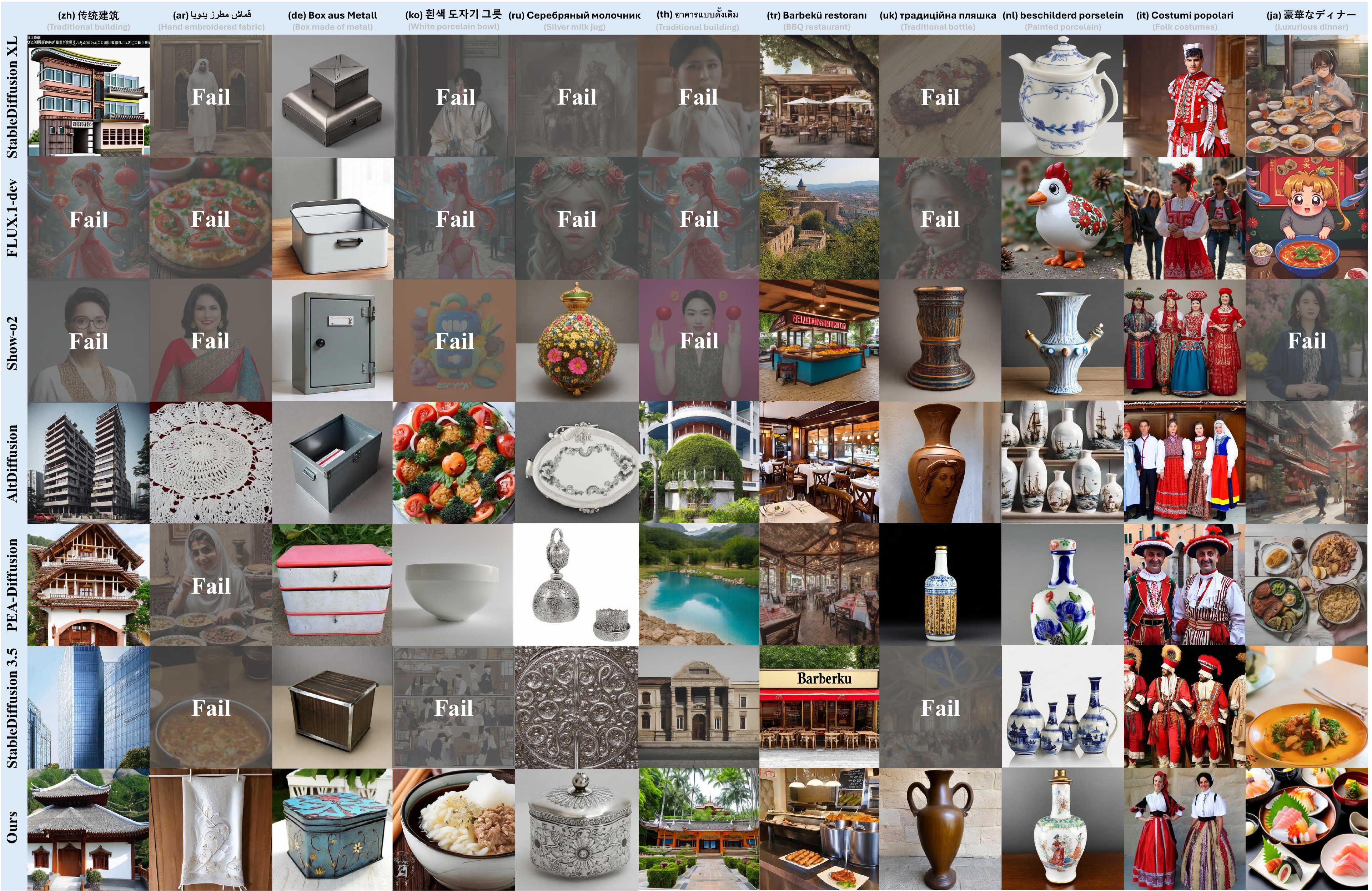}
\caption{\textbf{Qualitative comparison of generation results.}  Our approach generates images that are more culturally appropriate.}
\vspace{-0.4cm}
\label{fig:dingxing}
\end{figure*}

\section{Experiments And Analysis}

\subsection{Implementation Details}

\noindent\textbf{Metrics.} Following PEA-Diffusion~\cite{ma2024pea}, we employ CLIPScore~\cite{hessel2021clipscore} and ImageReward~\cite{xu2023imagereward} to evaluate text-image alignment and perceptual preference. Concurrently, we employ LPIPS~\cite{zhang2018unreasonable} to measure perceptual similarity and reconstruction fidelity, while introducing CultureVQA as a novel metric for cultural recognition.

\noindent\textbf{Hyperparameters.} In Fine-Tuned Layer Enhancer method, we train for 2{,}000 steps under mixed precision using AdamW (learning rate \(5\times10^{-5}\)) with a batch size of \(1\). For the zero-training variant, we set \(\lambda=6\). All experiments are conducted on a single NVIDIA A6000 GPU.

\subsection{Compare with SOTA Methods}
We conduct a quantitative comparison between our zero-training and fine-tuned built on PEA-Diffusion, and several state-of-the-art (SOTA) methods, including AltDiffusion~\cite{ye2024altdiffusion}, PEA-Diffusion~\cite{ma2024pea}, StableDiffusion XL~\cite{podell2023sdxl}, StableDiffusion 3.5~\cite{esser2024scaling}, FLUX.1-dev~\cite{labs2025flux1kontextflowmatching}, and Show-o2~\cite{xie2025show}. This evaluation measures semantic alignment and visual fidelity, providing a comprehensive benchmark for cross-cultural image generation.

\noindent\textbf{Quantitative Results.} To evaluate whether our cultural alignment strategies enhance cultural understanding without compromising semantic or visual quality, we designed a quantitative comparison experiment (Table~\ref{tab:quantitative}). The results show that our fine-tuned model achieves a CultureVQA score of 36.63, significantly outperforming AltDiffusion (23.05) and PEA-Diffusion (21.65). The zero-training variant attains the highest CLIPScore (0.291) and ImageReward (0.33), while maintaining a competitive LPIPS. Therefore, our method achieves simultaneous gains in cultural understanding, text-image consistency, and perceptual quality, verifying that interpretable cultural alignment can be realized without sacrificing visual fidelity.

\noindent\textbf{Qualitative Results.} To test whether our method can recover cultural features under “noun-only” prompts without losing semantic consistency, we conducted qualitative comparisons with several mainstream models (Figure~\ref{fig:dingxing}). The results show that our approach consistently generates images that reflect the target region’s cultural characteristics while preserving semantic accuracy, whereas other models often regress to culture-neutral generic prototypes. For example, under prompts such as “\emph{Box made of metal}” or “\emph{Silver milk jug}”, our method produces culturally grounded depictions, while PEA-Diffusion and AltDiffusion generate correct objects but lack regional style; StableDiffusion XL/3.5 and FLUX.1-dev tend toward neutral aesthetics, and Show-o2 exhibits both cultural inconsistency and semantic drift. Overall, our method balances cultural and semantic coherence while maintaining high image.

\begin{figure}[t]
\centering
\includegraphics[width=0.9\linewidth]{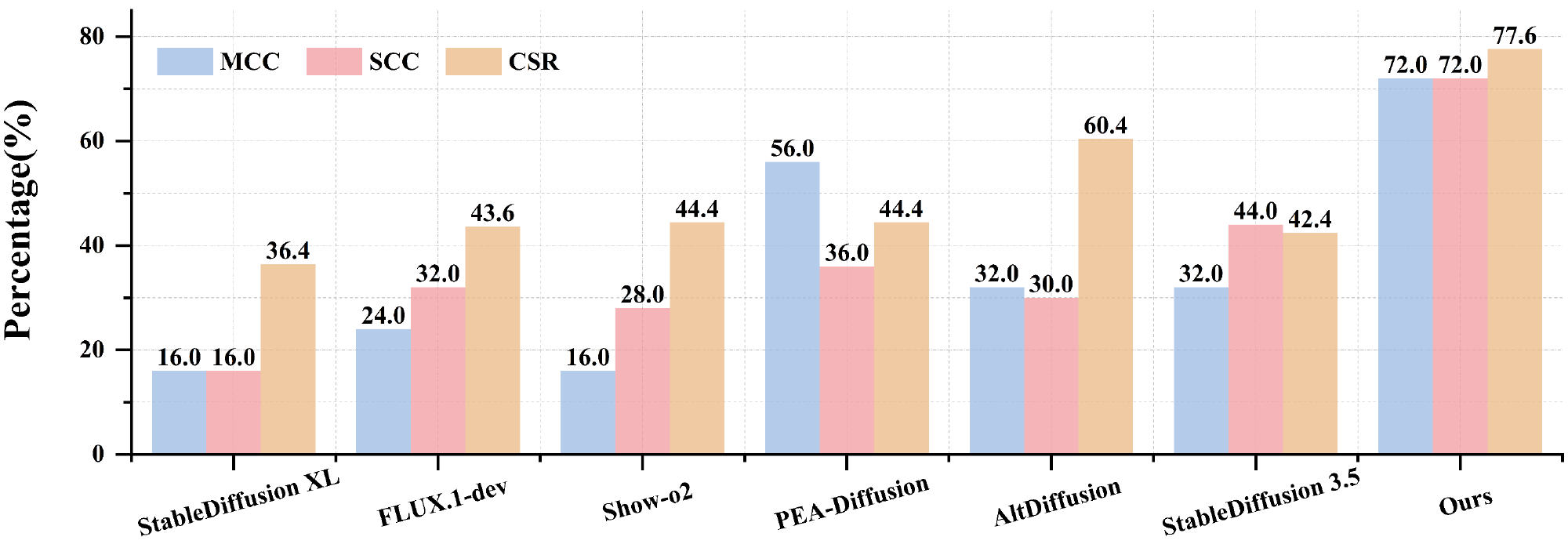}
\caption{\textbf{User Study.} Evaluated using MCC, SCC, and CSR metrics, where higher scores indicate greater perceived realism and user preference.}
\vspace{-0.3cm}
\label{fig:user}
\end{figure}

\begin{figure}[t]
\centering
\includegraphics[width=0.9\linewidth]{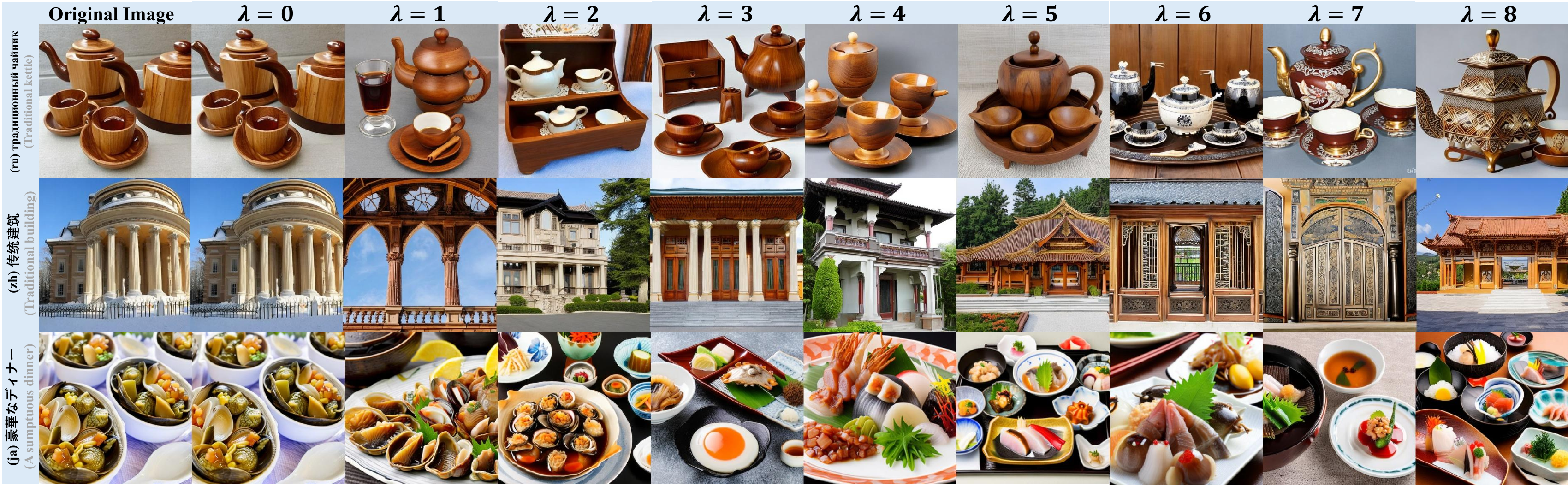}
\vspace{-0.3cm}
\caption{\textbf{Hyperparameter results.} The effect of cultural enrichment varies under different $\lambda$ values.}
\vspace{-0.5cm}
\label{fig:lambda}
\end{figure}

\noindent\textbf{User Study.} Building on the quantitative and qualitative results, we conducted a human-centered user study on the CultureBench platform with 50 experts in cultural studies. Participants evaluated cultural perception using three metrics: MCC (Multi-Choice Culture), SCC (Single-Choice Culture), and CSR (Cultural Semantic Relevance, 1–5 scale). Higher scores indicate stronger cultural alignment and preference. As shown in Figure~\ref{fig:user}, our method outperforms all baselines across all three metrics.
Notably, the human-rated CSR score reaches 77.6, significantly higher than the second-best score of 60.4, highlighting a clear advantage in cultural semantic fidelity. This human-evaluated CSR metric further strengthens our evaluation pipeline.
Overall, these results confirm that our method can generate culturally aligned content accurately and consistently.

\subsection{Hyperparameter Sensitivity Analysis}
\begin{figure}[t]
\centering
\includegraphics[width=0.85\linewidth]{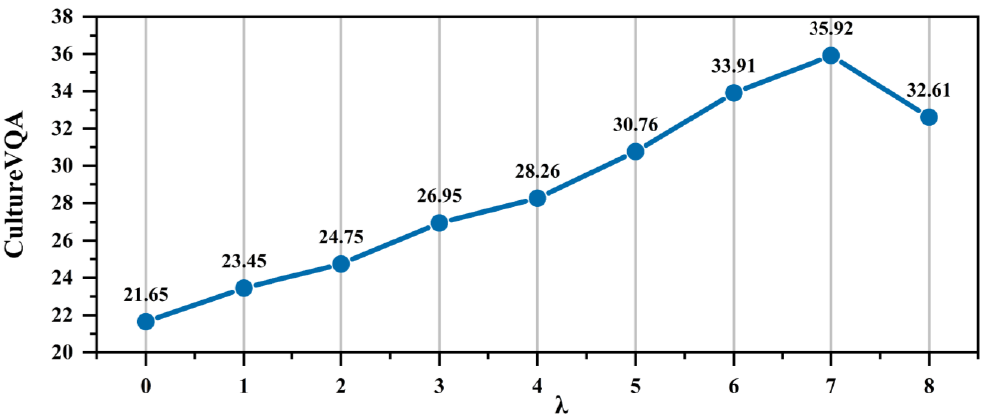}
\vspace{-0.3cm}
\caption{\textbf{Hyperparameter results.} Performance variations of CultureVQA under different $\lambda$ values.}
\vspace{-0.2cm}
\label{fig:lambda_value}
\end{figure}

To investigate the impact of hyperparameter $\lambda$ on the cultural attributes of generated images in zero-shot training, we compared results under different $\lambda$ values while keeping model parameters and the ``noun-only’’ prompt constant (see Figure~\ref{fig:lambda} and Figure~\ref{fig:lambda_value}). When $\lambda=0$, the output perfectly matched the original image. As $\lambda$ increases, the images progressively align with the prototypical features of the target culture, accompanied by a corresponding rise in CultureVQA scores. A peak is reached at $\lambda=7$ (35.92), followed by a slight decline at $\lambda=8$ (32.61). This indicates that $\lambda$ effectively modulates the intensity of cultural consistency, though excessively high values may induce overfitting and marginally degrade metric performance. Therefore, we select $\lambda=7$.

\renewcommand{\arraystretch}{1.05}
\begin{table}[t]
\centering
\caption{\textbf{Ablation studies on CultureBench.} On the CultureBench test subset, we conducted ablation analyses of two methods under both zero-training and fine-tuned settings. The best performance is marked in \textbf{bold}.}
\vspace{-0.3cm}
\adjustbox{max width=\linewidth}{
\begin{tabular}{clc}
\hline
\textbf{Model} & \textbf{Method} & \textbf{CultureVQA $\uparrow$} \\
\hline
\multirow{5}{*}{AltDiffusion~\cite{ye2024altdiffusion}} 
& \cellcolor{gray!15} \textit{w/o} Ours & \cellcolor{gray!15} 23.05 \\
\cline{2-3}
& \ \ \textit{w/} Random (Zero-Training) & 20.38 \ \scriptsize{\color{ForestGreen}{(-2.67)}} \\
& \ \ \textit{w/} Ours (Zero-Training) & \textbf{30.06} \ \scriptsize{\color{darkred}{(+7.01)}} \\
\cline{2-3}
& \ \ \textit{w/} Random (Fine-Tuned) & 21.04 \ \scriptsize{\color{ForestGreen}{(-2.01)}}  \\
& \ \ \textit{w/} Ours (Fine-Tuned) & \textbf{32.66} \ \scriptsize{\color{darkred}{(+9.61)}} \\
\hline
\multirow{5}{*}{PEA-Diffusion~\cite{ma2024pea}} 
& \cellcolor{gray!15} \textit{w/o} Ours & \cellcolor{gray!15} 21.65 \\
\cline{2-3}
& \ \ \textit{w/} Random (Zero-Training) & 21.04 \ \scriptsize{\color{ForestGreen}{(-0.61)}} \\
& \ \ \textit{w/} Ours (Zero-Training) & \textbf{33.91} \ \scriptsize{\color{darkred}{(+12.26)}} \\
\cline{2-3}
& \ \ \textit{w/} Random (Fine-Tuned) & 22.34 \ \scriptsize{\color{darkred}{(+0.69)}} \\
& \ \ \textit{w/} Ours (Fine-Tuned) & \textbf{36.63} \ \scriptsize{\color{darkred}{(+14.98)}} \\
\hline
\label{tab:ab}
\end{tabular}
}
\vspace{-0.8cm}
\end{table}

\subsection{Ablation Study}

To evaluate the effectiveness of our proposed components and ensure that the performance gains are not caused by random enhancement, we conducted ablation studies on the CultureBench (Table~\ref{tab:ab}). Both the zero-training and fine-tuned variants exhibit clear improvements: under the zero-training setting, CultureVQA increases by 7.01 on AltDiffusion and 12.26 on PEA-Diffusion; after fine-tuning, the gains further rise to 9.61 and 14.98, respectively. Random activation or random fine-tuning yields only minimal or even negative improvements. These results confirm that the performance gains stem from the targeted activation of culturally sensitive neurons. Furthermore, both AltDiffusion and PEA-Diffusion exhibit consistent performance improvements, further validating the universality and cross-architecture effectiveness of our neuron detection and enhancement framework.

\section{Conclusion}
In this work, we reveal that multilingual T2I models often yield culturally neutral or English-biased images not due to missing knowledge, but due to weak activation of culture-sensitive representations. By localizing culture signals to a small neuron set in a few layers, we introduce two lightweight remedies: inference-time cultural activation and layer-targeted cultural enhancement. On CultureBench, both strategies consistently improve cross-lingual cultural consistency while preserving fidelity and diversity. This points toward practical, controllable cultural alignment for inclusive generative models.

{
    \small
    \bibliographystyle{ieeenat_fullname}
    \bibliography{main}
}

\twocolumn[{%
\renewcommand\twocolumn[1][]{#1}%
\maketitlesupplementary
\begin{center}
    \newcommand{\teaserwidth}{\textwidth}
    \centerline{
        \includegraphics[width=1.0\teaserwidth,clip]{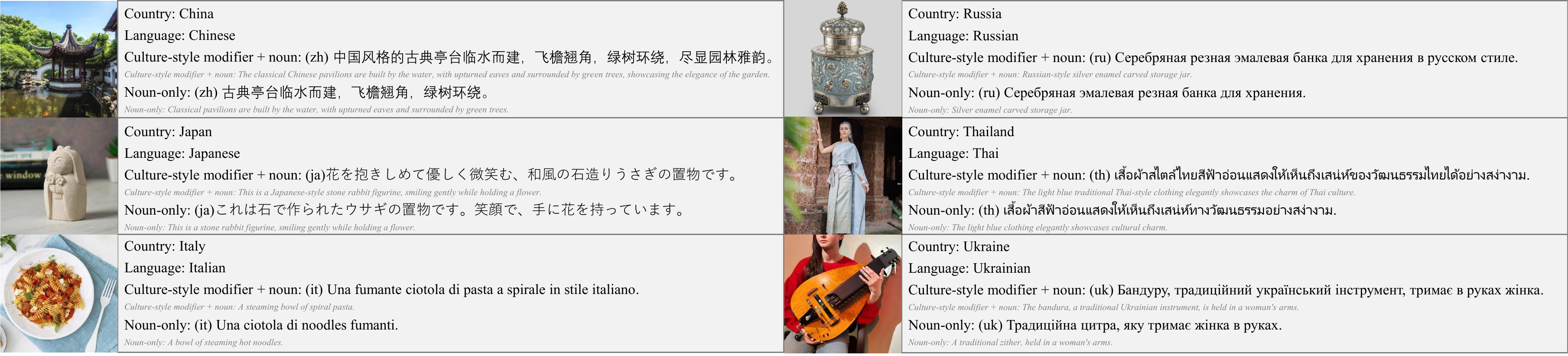}
    }
    \vspace{-0.07in}
    \captionof{figure}{{\bf Examples of the geographical and cultural composition of the CultureBench dataset.}
    }
    \vspace{-0.15in}
\label{fig:detail_data}
\end{center}%
}]

\appendix

The appendices provide additional details that support and extend the main paper. Appendix~\ref{Detail_Dataset} describes the dataset's geographic composition, evaluation protocol, and explains the selection of specific countries.
Appendix~\ref{gap} further validates the conjecture presented in the main text.
Appendix~\ref{more_exp} presents further experimental results and ablation studies.
Appendix~\ref{detail_user} details the user study's design and conduct.
Appendix~\ref{more_res} provides additional visualization results for PEA-Diffusion~\cite{ma2024pea} and AltDiffusion~\cite{ye2024altdiffusion} under both methods.
Appendix~\ref{diss} addresses common issues.
Appendix~\ref{lim} covers the limitations of our work.
Appendix~\ref{ethic} reflects on the ethical considerations of this research.

\section{Details of CultureBench Dataset}
\label{Detail_Dataset}
This study presents the dataset's geographical composition within a broad cultural classification framework. It encompasses major cultural spheres such as the Arab world, East Asia, continental Europe, Latin America, and parts of Africa. This approach reflects the macro-sociological context of the data sources. It does not represent an essentialist or homogenized understanding of cultures. Cultural spheres are delineated by principal nations and languages, including Arabic, Chinese, Japanese, Korean, Thai, German, Russian, Italian, Dutch, Polish, Turkish, Ukrainian, Spanish, Portuguese, and French in parts of Africa. Figure~\ref{fig:detail_data} illustrates an exemplar composition of this study's dataset.

\noindent\textbf{Details of Data Collection.} 
During data collection, we used Google Image Search~\footnote{\url{https://images.google.com}} and complementary web resources. These included Wikipedia~\footnote{\url{https://www.wikipedia.org}} and other public search platforms. We obtained publicly available images, setting keywords based on various cultural contexts to cover diverse scenarios and object types. After automated extraction, images undergo preliminary screening. This eliminates low-quality, semantically irrelevant, or copyright-infringing samples. A cross-cultural expert team then manually reviews images, prioritizing the exclusion of those conveying stereotypes or biases. In addition, these experts define and validate expert-curated cultural subdivisions within each broad cultural group. These include regional styles, ethnic or indigenous traditions, and locally distinctive visual elements. The team uses these subdivisions to guide manual filtering and annotation. This ensures an accurate representation of the intended cultural context without incorporating visual elements from other cultural backgrounds. As a result, CultureBench captures both macro-level cultural identity and finer-grained intra-cultural diversity. Through this process, we have enhanced the cultural accuracy and fairness of our cross-cultural visual data while ensuring its diversity.

\section{Further Evidence on the Culture Gap}
\label{gap}
\begin{figure*}[t]
\centering
\includegraphics[width=0.9\linewidth]{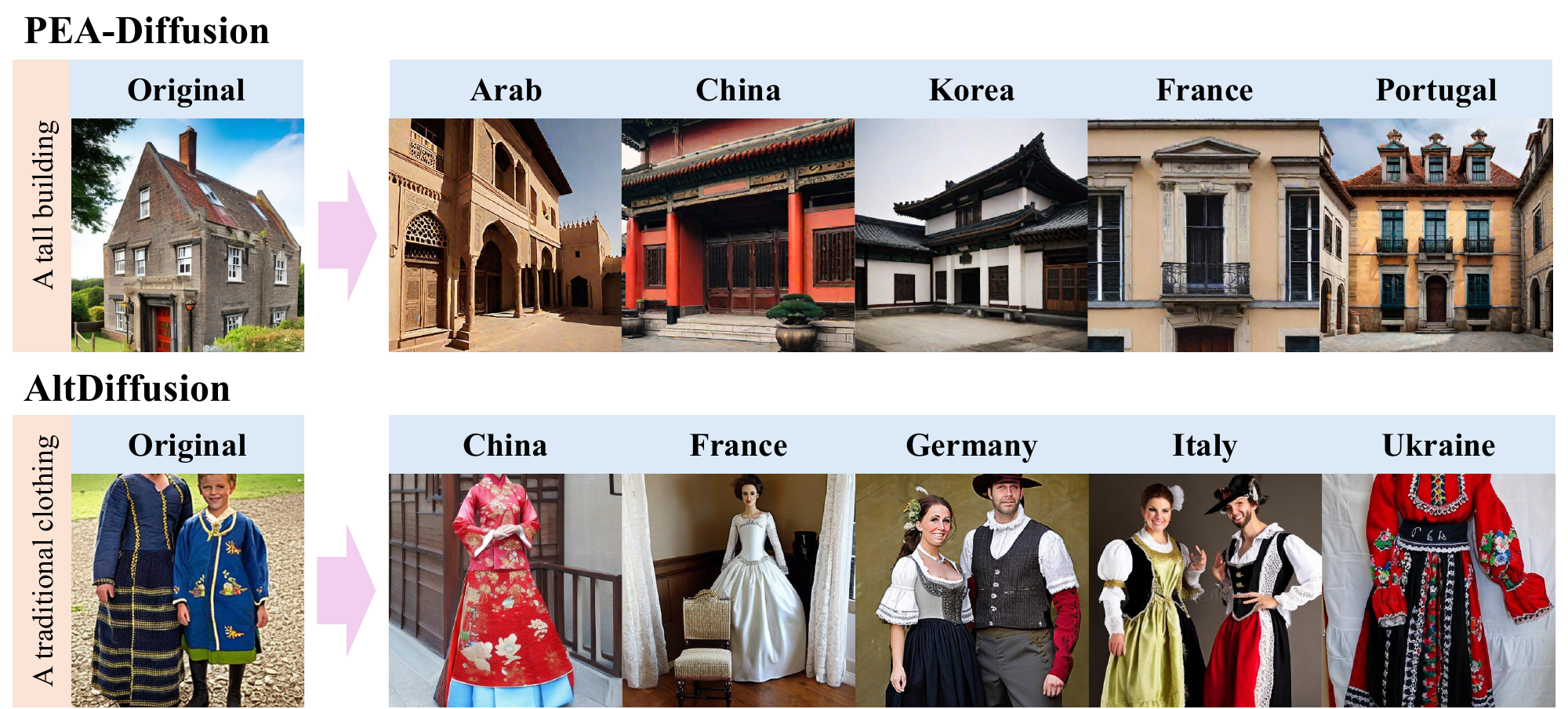}
\vspace{-0.3cm}
\caption{\textbf{Further evidence on the culture gap.} For a fixed English prompt and identical sampling settings, activating different culture-neuron sets steers both PEA-Diffusion (top) and AltDiffusion (bottom) toward distinct cultural styles.}
\label{fig:ganyu}
\end{figure*}

To test whether the cultural gap arises from under-activation rather than knowledge absence, we conduct a unified controlled experiment (Figure.~\ref{fig:ganyu}). Using a culture-neutral English prompt and fixing all stochastic factors (e.g., seed and sampler), we generate images while selectively activating different culture-neuron sets identified in previous analyses. Despite identical prompts, activating China, Japan, Italy, Germany, or Ukraine neuron sets yields clearly distinct cultural styles, whereas the baseline remains culturally neutral. This controlled intervention demonstrates that the model already encodes rich culture-specific representations, and that the failure stems primarily from insufficient activation during generation rather than missing cultural knowledge.

\begin{table}[t]
    \renewcommand{\arraystretch}{1.2}
    \centering
    \caption{\textbf{Comparison of accuracy between CultureVQA and human experts on the CultureBench dataset.}}
    \begin{tabular}{l|c}
        \hline
        \hline
        Model / Group & Accuracy \\
        \hline
        Human Experts (Avg.) & 94.18\% \\
        \textbf{CultureVQA (Ours)} & 91.57\% \\
        \hline
        \hline
    \end{tabular}
    \label{tab:culturevqa_vs_experts}
\end{table}

\section{More Experiments}
\label{more_exp}

\subsection{Reliability of CultureVQA}
\label{Reliability}
To assess the consistency of CultureVQA with human subjective cognition, we invited 30 domain experts to participate in comparative experiments using the CultureBench dataset. Each question contained four real-world images, and only one matched the specified cultural element. CultureVQA selected one image per question, and the experts made their choices under the same conditions. We then calculated and compared the accuracy rates for both CultureVQA and the experts. As shown in Table~\ref{tab:culturevqa_vs_experts}, CultureVQA achieved an accuracy of 91.57\%, while human experts reached an average accuracy of 94.18\%, resulting in a gap of only 2.61 percentage points. This relatively small difference indicates that CultureVQA’s performance closely approaches that of human experts, demonstrating high consistency with human cultural recognition in this task.

\subsection{Cultural Probing Universal Type}
For clarity in the main text, only the detection results from PEA-Diffusion are presented. This section additionally demonstrates the culture-sensitive layers and neurons associated with AltDiffusion for comparison purposes.

\begin{figure}[t]
\centering
\includegraphics[width=0.9\linewidth]{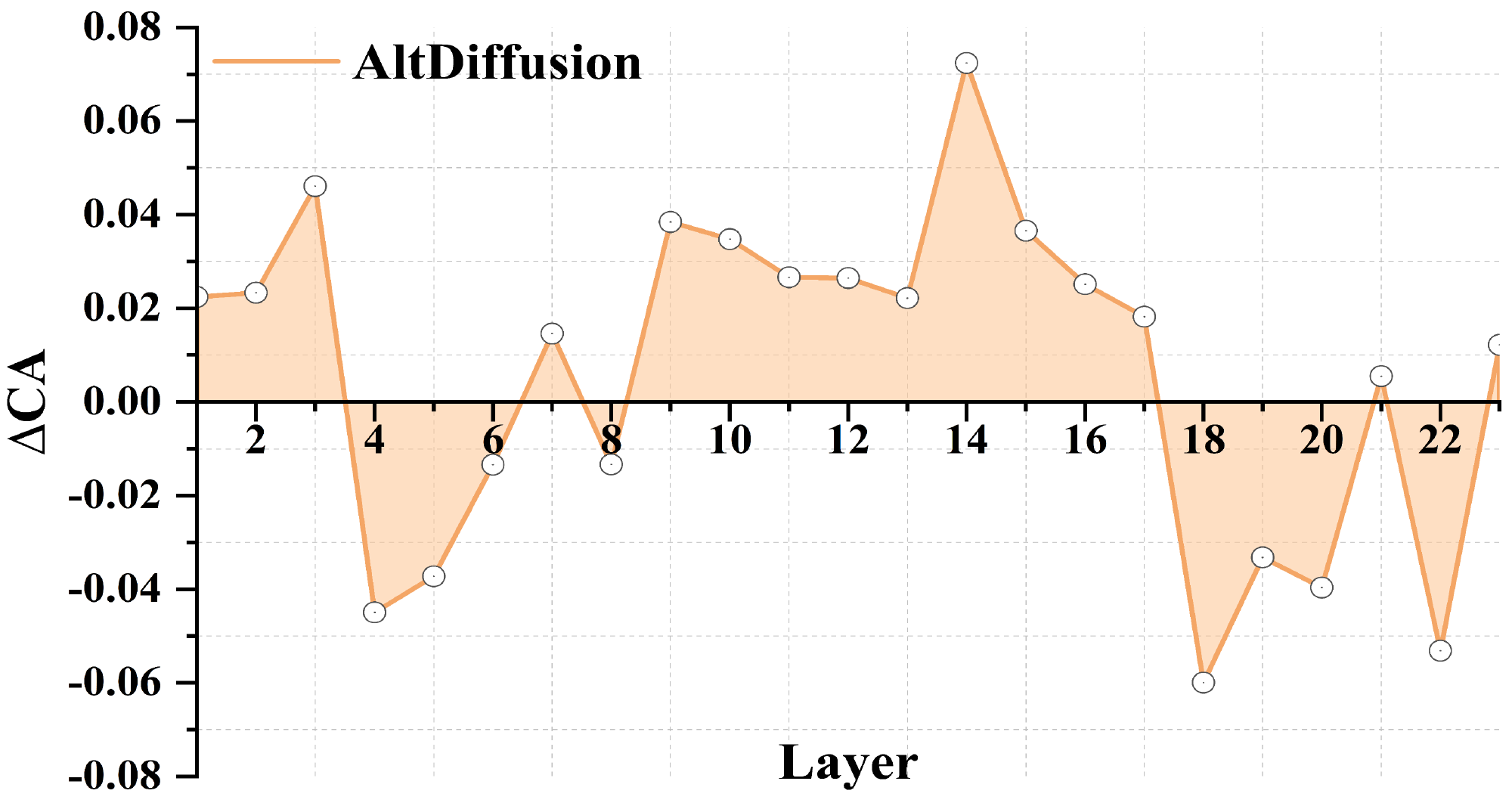}
\caption{\textbf{AltDiffusion cultural sensitivity.} $\Delta$CA peaks layer 14. Therefore, layer 14 is culturally sensitive.}
\vspace{-0.3cm}
\label{fig:alt_layer}
\end{figure}

\begin{figure}[t]
\centering
\includegraphics[width=1\linewidth]{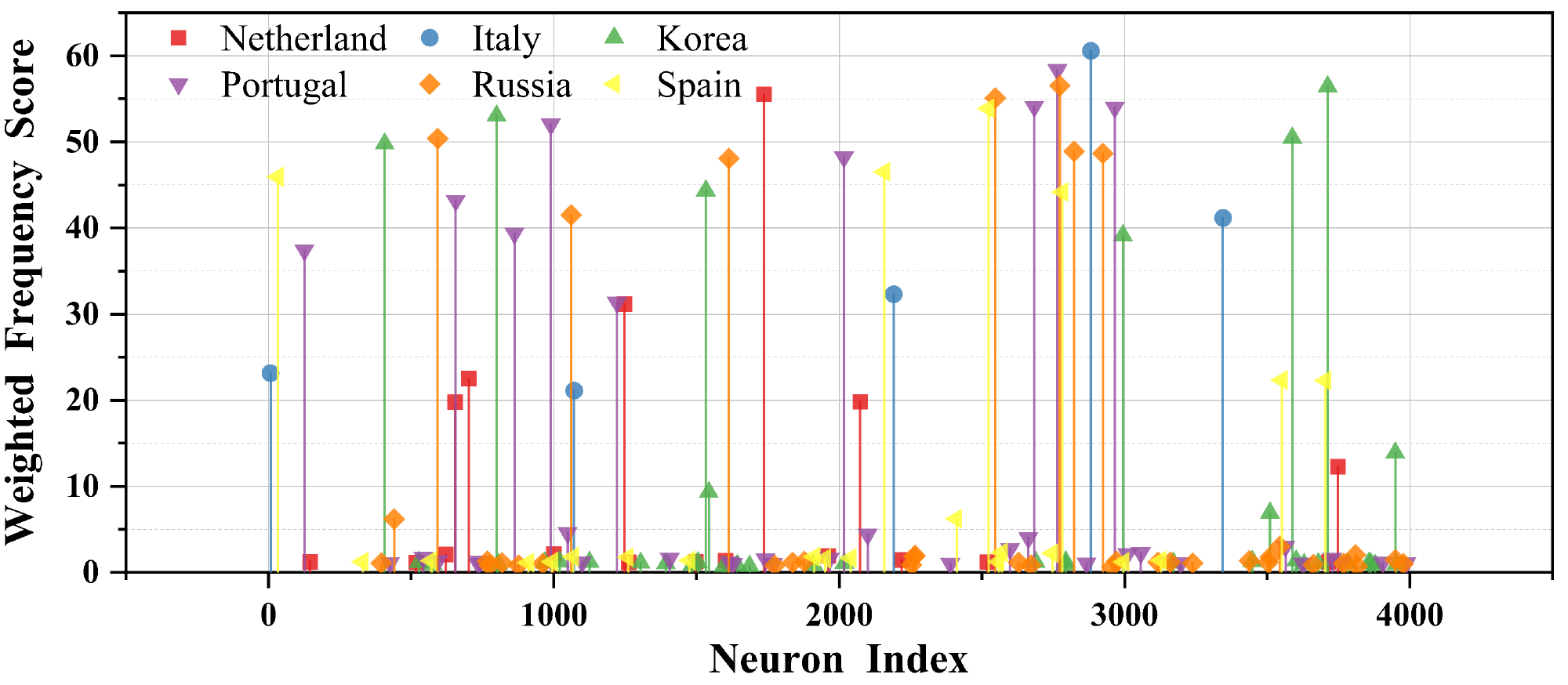}
\caption{\textbf{AltDiffusion neuronal detection result.} The weighted frequency scores show only a few salient peaks per culture, indicating culture-specific neurons.}
\vspace{-0.5cm}
\label{fig:alt_ne}
\end{figure}

\noindent\textbf{Culture Layer Detection.}
To test the generality of our detection method beyond PEA-Diffusion, we also analyze AltDiffusion. Following the main paper's probing procedure, we create paired prompts (``culture-style modifier + noun'' and ``noun-only''), annotate modifier and noun token groups, and extract cross-attention maps from all layers. We then compare aggregated attention from cultural modifiers to their paired nouns, yielding layerwise cultural-attention contrast curves for AltDiffusion (Figure~\ref{fig:alt_layer}).

\noindent\textbf{Culture Neuron Detection.}
We employ the same methodology as in the main text to localize culturally sensitive neurons within the AltDiffusion culture layer. As illustrated in Figure~\ref{fig:alt_ne}, we present culturally sensitive neurons across six distinct cultural contexts. This demonstrates our approach's robust capability to detect culturally sensitive neurons across varied architectural frameworks.

\subsection{Cultural Validation in AltDiffusion}
To analyze the accuracy of culture neurons in AltDiffusion, we validate these neurons for each cultural group. For all 15 cultural groups, we report CultureVQA scores in three controlled settings: (1) Baseline (no masking), (2) Masked Top-K Neurons (masking the Top-K identified culture-sensitive neurons), and (3) Random Mask (masking the same number of neurons at random). Table~\ref{tab:alt_cacc_ablation} shows that masking the Top-K neurons causes a substantial drop in CultureVQA performance. The mean score falls by 32.50 points compared to the AltDiffusion baseline. In contrast, randomly masking the same number of neurons reduces the mean score by just 2.09 points. This negligible drop stays close to the baseline. These results indicate that the culture-sensitive neurons we identify in AltDiffusion concentrate cultural information and capture cultural representations.

\renewcommand{\arraystretch}{1}
\begin{table}[t]
\centering
\caption{\textbf{Validating Culture-Sensitive Neuron Detection in AltDiffusion.} Neuronal accuracy on the test subset with “cultural style modifier + noun” prompts.}
\label{tab:alt_cacc_ablation}
\adjustbox{max width=\linewidth}{
\begin{tabular}{l|c}
\hline
Method & CultureVQA $\uparrow$ \\
\hline
\rowcolor{gray!20}AltDiffusion~\cite{ye2024altdiffusion} & 44.54 \\
+ Masked Top-K Neurons & 12.04 \scriptsize{\color{darkred}{(-32.50)}} \\
+ Masked Random Neurons & 42.45 \scriptsize{\color{darkred}{(-2.09)}} \\
\hline
\end{tabular}
}
\end{table}

\begin{figure}[t]
\centering
\includegraphics[width=1\linewidth]{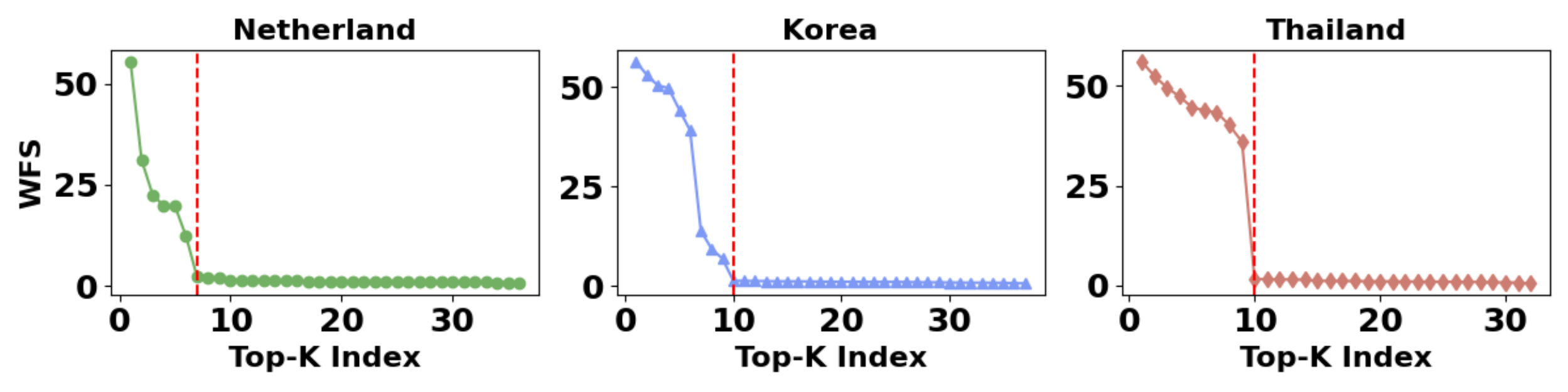}
\vspace{-0.3cm}
\caption{\textbf{Selection of Threshold $K$.} The red dashed vertical line indicates the chosen threshold $K$. Neurons to the left of the line correspond to the selected Top-$K$ culture-sensitive neurons, while the scores to the right fall below the cutoff and are discarded.}
\label{fig:K_selection}
\end{figure}

\subsection{Selection of Threshold $K$}
To clarify how we choose the threshold $K$, we visualize the weighted frequency scores (WFS) of candidate neurons and inspect their response patterns. The main factor guiding the choice of $K$ is the presence of a small subset of neurons that exhibit prominently higher responses than the rest. As shown in Figure~\ref{fig:K_selection}, we plot the WFS curves for three different cultural groups; in each case, a few leading neurons form clear peaks, followed by a rapid decay. We set $K$ at the elbow before this sharp drop, so that neurons with salient high responses are retained, while the long tail of weakly responsive neurons is discarded.

\subsection{Details of SAE}
we set the sparsity coefficient to \(\alpha = \frac{1}{32}\).  We adopt a Top-$K$ SAE with a hidden layer dimension of 4096. The model is optimized with AdamW using a learning rate of 0.0004 and a constant learning-rate schedule without warmup. We use an MSE reconstruction loss to encourage faithful feature reconstruction.

\subsection{Cross-Domain Generalization Results}
\label{cross}
We evaluated our method's generalization using quantitative and qualitative tests on cue words not present in the CultureBench dataset. We used 100 out-of-domain captions to ensure the evaluation reflects true and reliable out-of-distribution performance.

\renewcommand{\arraystretch}{1}
\begin{table}[t]
\centering
\caption{\textbf{Quantitative analysis across domains.} Generate using 100 captions from outside the domain and calculate the performance of CultureVQA. \textbf{Bold} rows mark the highest performance within each baseline group.}
\vspace{-0.1cm}
\adjustbox{max width=\linewidth}{
\begin{tabular}{lc}
\hline
\textbf{Method} & \textbf{CultureVQA $\uparrow$} \\
\hline
StableDiffusion XL~\cite{podell2023sdxl} & 10.00 \\
FLUX.1-dev~\cite{labs2025flux1kontextflowmatching} & 17.00 \\
Show-o2~\cite{xie2025show} & 15.00 \\
PEA-Diffusion~\cite{ma2024pea} & 15.00 \\
AltDiffusion~\cite{ye2024altdiffusion} & 15.00 \\
StableDiffusion 3.5~\cite{esser2024scaling} & 17.00 \\
\hline
\emph{\small{Baseline: PEA-Diffusion}} & \\
\rowcolor{lightgray} \textbf{Ours (Zero-Training)} & 22.00 \scriptsize{\color{darkred}{(+7.00)}} \\
\rowcolor{lightgray} \textbf{Ours (Fine-Tuned)} & \textbf{35.00} \scriptsize{\color{darkred}{(+20.00)}}  \\
\emph{\small{Baseline: AltDiffusion}} & \\
\rowcolor{lightgray} \textbf{Ours (Zero-Training)} & 20.00 \scriptsize{\color{darkred}{(+5.00)}}  \\
\rowcolor{lightgray} \textbf{Ours (Fine-Tuned)} & \textbf{32.00} \scriptsize{\color{darkred}{(+17.00)}}  \\
\hline
\end{tabular}
}
\vspace{-0.3cm}
\label{tab:cross_quan}
\end{table}

\begin{figure*}[t]
\centering
\includegraphics[width=1\linewidth]{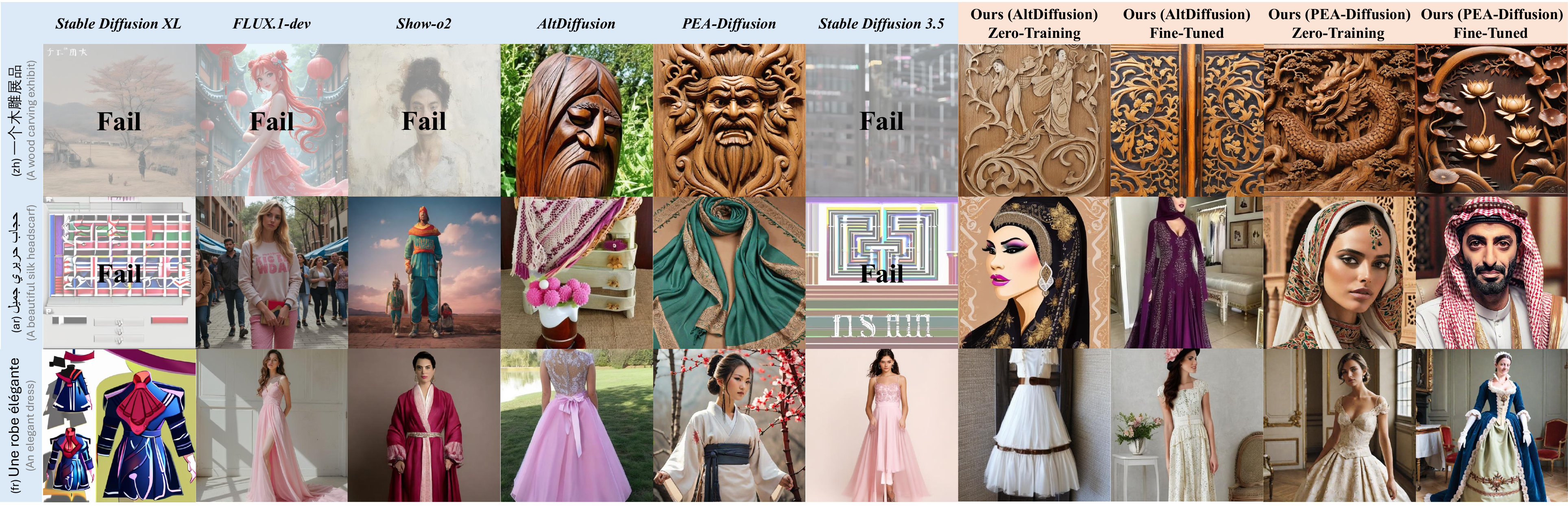}
\vspace{-0.4cm}
\caption{\textbf{Cross-domain qualitative experiments.} Our approach generates images that are more culturally appropriate.}
\label{fig:cross_dingxing}
\end{figure*}

\noindent\textbf{Quantitative Results.} Table~\ref{tab:cross_quan} shows that our method consistently improves cross-domain cultural accuracy for both PEA-Diffusion and AltDiffusion. Under the zero-training setting, our approach yields gains of +7.00 and +5.00 CultureVQA points over PEA-Diffusion and AltDiffusion, respectively, outperforming all existing off-the-shelf models. When fine-tuned, the improvements increase to +20.00 and +17.00 points for PEA-Diffusion and AltDiffusion, respectively, with our method achieving the highest accuracy within each baseline group. These results show that the culture-sensitive neurons identified by our method provide a substantial, transferable cultural prior, enabling robust generalization to captions outside the training distribution.

\noindent\textbf{Qualitative Results.} As illustrated in Figure~\ref{fig:cross_dingxing}, when prompted with the Chinese instruction ``\emph{A wood carving exhibit}'', most models exhibited semantic comprehension errors. Their outputs were inconsistent with the prompt's meaning. Our approach, however, maintained both semantic and cultural coherence. In contrast to Arabic instructions like ``\emph{A beautiful silk headscarf}'' or French prompts such as ``\emph{An elegant dress}'', our approach offers greater precision. It captures key textual details regarding fabric texture, style, and aesthetic quality. It also adeptly integrates regional cultural symbols and aesthetic preferences in color coordination, pattern design, and figure representation. This enables the generation of images better suited to local cultural contexts. Taken together, these qualitative findings indicate that our approach is not only more robust and expressive but also more culturally sensitive across languages and regions. It substantially improves cross-linguistic and cross-cultural text-image alignment and image generation, thereby further validating its effectiveness and broad applicability in real-world, culturally diverse scenarios.

\begin{figure*}[t]
\centering
\includegraphics[width=0.95\linewidth]{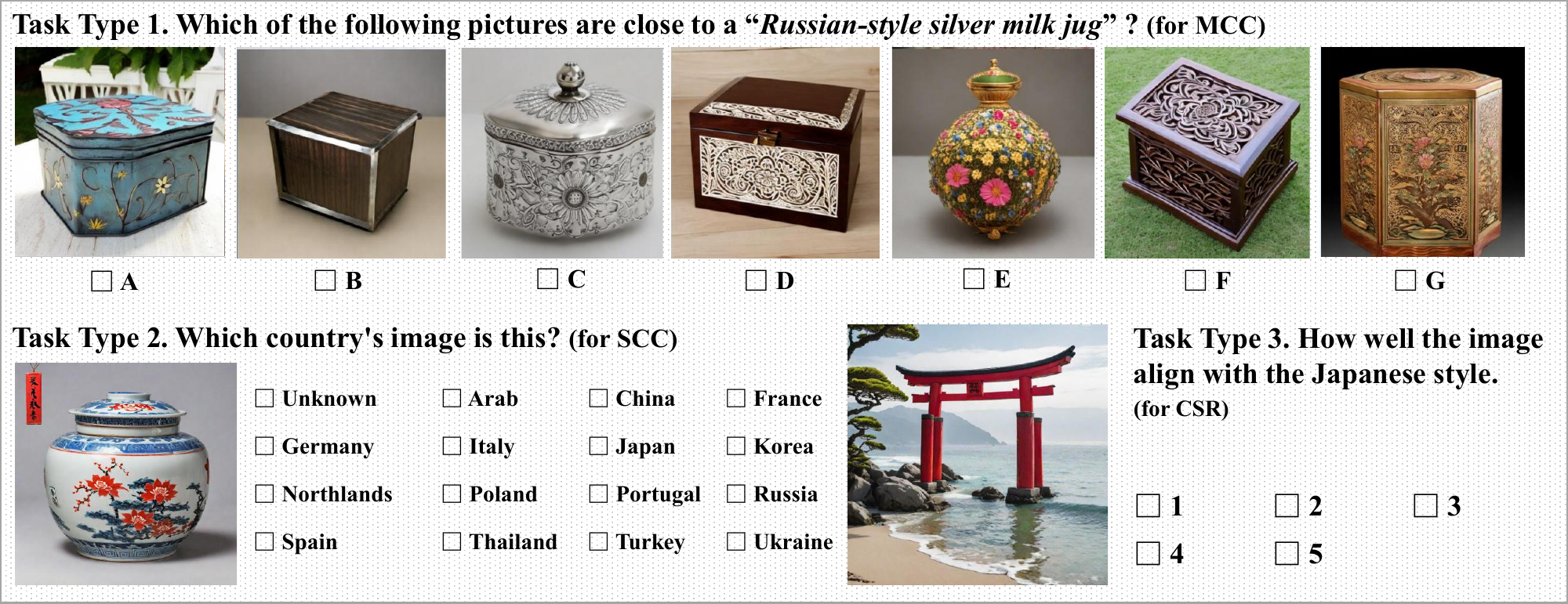}
\caption{\textbf{Example interface of the user study comprising three task types.} MCC (top), SCC (bottom-left), and CSR (bottom-right).}
\label{fig:user_detail}
\end{figure*}

\begin{figure*}[t]
\centering
\includegraphics[width=0.95\linewidth]{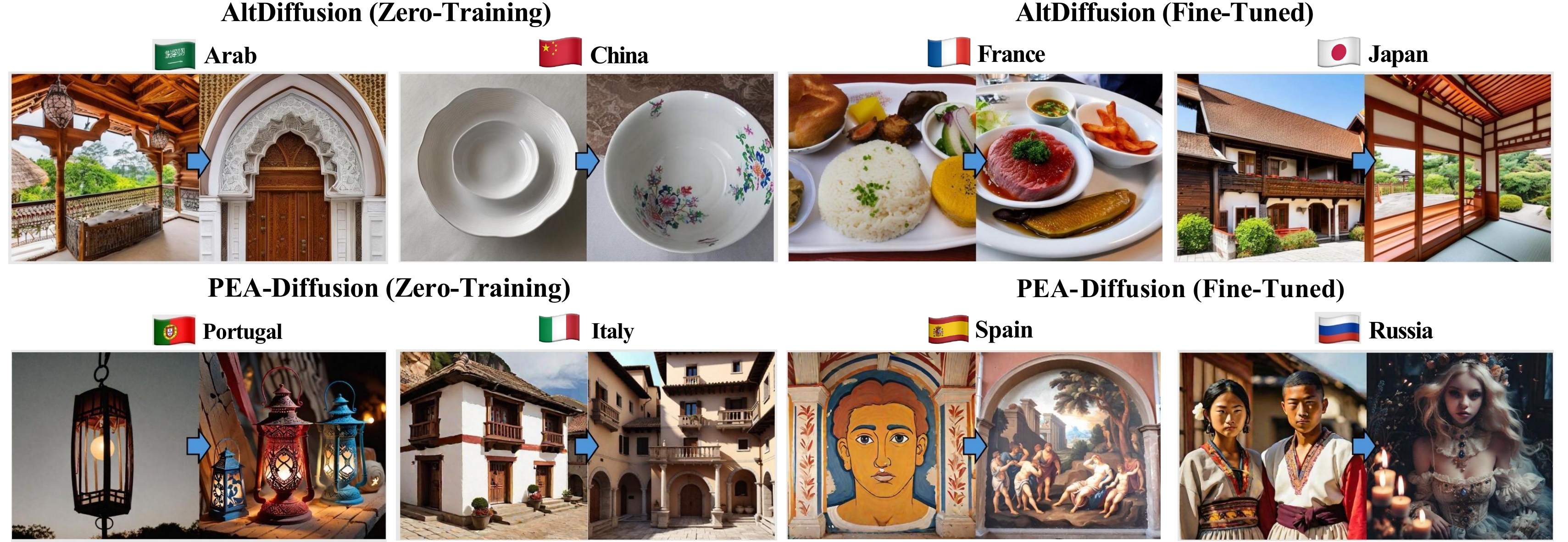}
\vspace{-0.2cm}
\caption{\textbf{More Results.} Further examples of results generated by AltDiffusion and PEA-Diffusion under different methods.}
\vspace{-0.5cm}
\label{fig:more}
\end{figure*}

\section{Detail of User Study}
\label{detail_user}
As shown in Figure~\ref{fig:user_detail}, we conducted a controlled user study to evaluate the perceptual effectiveness of this method. Fifty participants, aged 25-47 (23 female), completed three sequential perceptual tasks. Each task assessed the subjective performance of generated images, focusing on cultural alignment and preference. 

\noindent\textbf{Multi-Choice Culture.} During the Multi-Choice Culture (MCC) mission, we first used the noun-only prompt within CultureBench and fed it to multiple image generation models, obtaining numerous candidate images. Next, participants were asked to select all images they deemed representative of a given target culture. This selection process then measured each model's capacity to generate culturally coherent images without textual cultural cues. 

\noindent\textbf{Single-Choice Culture.} Single-Choice Culture (SCC) defines cultural perception as a forced-choice task of cultural identification. For each noun-only prompt, we generate an image using a specific model. Participants view the model-generated image and are asked to infer its most probable cultural category. We provide a fixed set of 16 options: 15 predefined cultural categories plus an additional 'Uncertain' option. Participants must select a single option based precisely on the image's visual content.

\noindent\textbf{Cultural Semantic Relevance.} Cultural Semantic Relevance (CSR) is employed to evaluate the subjective cultural semantic alignment of images within a given target culture. For each CSR item, we present an image generated by a specific model, explicitly informing participants of its stylistic origin. Participants are required to rate the image's alignment with the target culture within the context of that model's style, on a scale of 1 to 5.

\section{More Results}
\label{more_res}
To demonstrate the effectiveness of our approach more intuitively, we generated both an unenhanced baseline image and an enhanced image using the same prompt, while keeping the model's seed and other generative parameters fixed. The comparison between the two is shown in Figure~\ref{fig:more}.
The results show that the enhanced images more faithfully reflect the cultural context, including scene elements, character depictions, and fine textures. This leads to more accurate visual expressions of the target culture and further highlights the effectiveness of our approach in achieving cultural consistency.

\section{More Discussion}
\label{diss}

\noindent$\triangleright$ \textbf{\textit{Q1. What do we mean by “cultural consistency” in this work?}}
We define cultural consistency as the degree to which an image from a multilingual prompt shows statistically grounded, contextually appropriate cultural cues tied to the target language’s socio-cultural context, rather than merely literal semantic correctness. Our approach uses (i) a fixed language–region mapping, (ii) observable, moderate visual grounding (such as architecture, clothing, artifacts), and (iii) expert screening to remove stereotypical or inappropriate cues. These elements turn a vague idea into an operational objective that can be measured and optimized.

\noindent$\triangleright$ \textbf{\textit{Q2. Why is CultureBench designed as a medium-scale diagnostic benchmark rather than a larger dataset?}}
CultureBench is deliberately scaled to a moderate size, as it serves as a controlled diagnostic benchmark for assessing the cultural behaviour of multilingual text-to-image models rather than functioning as a pre-training corpus: Approximately 7.9k images, meticulously annotated across 15 language-culture regions, suffice to reveal systemic cultural omissions under ``noun-only'' prompts and support neural-level analysis, whilst ensuring the feasibility of high-quality, de-stereotyped annotation. We thus regard it as an extensible starting point rather than a comprehensive catalogue covering all cultures.

\noindent$\triangleright$ \textbf{\textit{Q3. Why do we rely on CultureVQA, and how reliable is it as an evaluation metric?}}
We adopt CultureVQA because it directly addresses the challenge of evaluating whether images accurately reflect the nuanced cultural attributes encoded in multilingual prompts. Unlike existing automatic metrics (e.g., FID, CLIPScore), which focus on pixel-level similarity or general correspondence, CultureVQA frames evaluation as a semantic recognition problem, allowing assessment of cultural correctness. By using a VQA-style cultural identification task, CultureVQA enables consistent, scalable, and multi-class evaluation across 15 cultural regions. Its reliability is supported by a human–model consistency study (Appendix~\ref{Reliability}).

\noindent$\triangleright$ \textbf{\textit{Q4. What evidence supports the hypothesis that cultural knowledge exists but is under-activated?}}
Our claim is posed as a hypothesis and supported by converging empirical evidence rather than a formal proof. First, explicit culture-style modifiers (e.g., “\emph{Italian architecture}”, “\emph{a person in traditional Chinese clothing}”) consistently trigger culturally grounded generations. In contrast, noun-only prompts collapse to neutral, culture-agnostic prototypes, indicating that the cultural capability is present but insufficiently activated. Second, masking the Top-K neurons identified by our probing method results in a sharp drop in CultureVQA performance, whereas masking the same number of random neurons has minimal effect, suggesting a causal relationship between these neurons and cultural semantics. Third, attention-based probing reveals a stable culture-sensitive layer in which ``culture-style modifier + noun'' and ``noun-only'' prompts yield distinct activation patterns. Together, these findings show that the model has cultural representations, but they are under-activated without explicit cues.

\noindent$\triangleright$ \textbf{\textit{Q5. Why are our neuron- and layer-level interventions reasonable, and do they over-claim causality?}}
Our neuron- and layer-level interventions are deliberately minimal and localized: both the zero-training neuron amplifier and the fine-tuned layer enhancer operate only within the culture-sensitive layer and within a small subspace identified by our probing method, without modifying the diffusion backbone. Adjusting these neurons reliably improves CultureVQA scores and human-perceived cultural fidelity, while leaving CLIPScore, ImageReward, LPIPS, and visual diversity essentially unchanged. This indicates that the intervention is targeted rather than disruptive. Importantly, we do not claim to establish definitive causal attributions for individual neurons. Instead, we frame our approach as a lightweight and interpretable control mechanism that leverages empirically responsive subspaces. It offers practical gains in cultural consistency while serving as a useful starting point for future, more formal causal analyses.

\noindent$\triangleright$ \textbf{\textit{Q6. Is testing only on CultureBench lacking external validation?}}
We agree that relying solely on CultureBench would limit external validation, which is why we include a cross-domain experiment using 100 out-of-distribution captions not appearing in CultureBench (Appendix~\ref{cross}). Our method still yields consistent gains in CultureVQA under both zero-training and fine-tuning in this out-of-domain setting, showing that the improvement is not tied to CultureBench’s specific prompts.

\noindent$\triangleright$ \textbf{\textit{Q7. This paper uses ChatGPT data to refine cultural products, but could bias and errors be introduced?}}
The use of ChatGPT-generated text may introduce biases; hence, we have implemented multiple dedicated measures in our data construction process to minimise these risks. Firstly, the ChatGPT content we utilise is strictly confined to ``culture-style modifier + noun''. All corresponding images are sourced exclusively from publicly available real-world reference materials and undergo multi-stage human expert review to eliminate stereotypes, semantic errors, and culturally inaccurate cues. Furthermore, the ChatGPT-generated modifiers themselves undergo expert filtering to ensure they do not contain stereotypical or inappropriate cultural descriptions.

\noindent$\triangleright$ \textbf{\textit{Q8. When switching generative models, is it necessary to re-execute the ``Cultural Sensitivity Layer and Cultural Neuron Detection'' step within our methodology?}}
Whether re-detection is required depends primarily on whether the text encoder of the new model has changed. If the new generative model continues to use the same text encoder as in our experiments, re-detection is generally unnecessary; if the text encoder has been altered, the detection steps must be re-executed.

\noindent$\triangleright$ \textbf{\textit{Q9. Mean square error (MSE) cannot assess cultural consistency, so is it appropriate to use it?}}
MSE is not in itself an effective metric for measuring cultural consistency, and we therefore did not employ it as an evaluation criterion. Within our framework, MSE serves as the training signal for specific layer enhancers, rather than the objective we use to gauge cultural consistency.

\noindent$\triangleright$ \textbf{\textit{Q10. What are the roles of $\epsilon$ and $\beta$?}}
$\epsilon$  is set to 0 by default, since activations greater than zero are treated as neuron firing. $\beta$ is a small stability constant added to the denominator to prevent it from becoming zero when a neuron has zero or very few activations.

\section{Limitation}
\label{lim}
CultureBench currently covers 15 cultural regions, but this still reflects only a limited portion of global cultural diversity. Because the benchmark is built from publicly available image resources, some cultures, especially those from low-resource or marginalized communities, remain underrepresented. We emphasize that the currently included cultural regions are carefully curated and remain valid, but they do not yet exhaust the diversity within each region or across countries. In future iterations, we intend to identify underrepresented cultural groups by partnering with relevant organizations, actively seeking additional image sources, and introducing finer intra-cultural subdivisions, thereby expanding the benchmark to more countries and territories for a more comprehensive and inclusive evaluation suite.

\section{Ethics Statement}
\label{ethic}
In this research, we acknowledge the potential misuse of image synthesis techniques, such as ours, for generating deceptive content and spreading disinformation, a serious concern we address explicitly. However, we also note the substantial progress made in detection and prevention mechanisms in this domain. Our framework supports critical research initiatives and encourages third-party oversight, aiming to strike a balance between technological advancement and security considerations. This balanced approach promotes responsible deployment while preserving the innovation potential.

\end{document}